%% file: emnlp2020.tex
\documentclass[11pt,a4paper]{article}
\usepackage[hyperref]{emnlp2020}
\usepackage{times}
\usepackage{latexsym}

\aclfinalcopy
\usepackage{microtype}

\usepackage[utf8]{inputenc} 
\usepackage[T1]{fontenc}    
\usepackage{url}            
\usepackage{booktabs}       
\usepackage{amsfonts}       
\usepackage{nicefrac}       
\usepackage{microtype}      

\usepackage{helvet}  
\usepackage{courier}  
\usepackage{adjustbox}
\usepackage{xcolor}
\usepackage{amsmath,amssymb}
\usepackage{bm} 

\usepackage{tabularx}

\usepackage{graphicx} 
\usepackage{subfig}
\graphicspath{{figures/}}
\usepackage{wrapfig}
\usepackage{algorithm}
\usepackage{algorithmic}

\usepackage{multirow}

\usepackage{breakcites}

\allowdisplaybreaks[1]

\input{mlVecMat.tex}
\ifx\proof\undefined
\newenvironment{proof}{\par\noindent{\bf Proof\ }}{\hfill\BlackBox\\[2mm]}
\fi

\ifx\theorem\undefined
\newtheorem{theorem}{Theorem}
\fi
\ifx\example\undefined
\newtheorem{example}{Example}
\fi
\ifx\lemma\undefined
\newtheorem{lemma}[theorem]{Lemma}
\fi

\ifx\corollary\undefined
\newtheorem{corollary}[theorem]{Corollary}
\fi

\ifx\assumption\undefined
\newtheorem{assumption}{Assumption}
\fi

\ifx\definition\undefined
\newtheorem{definition}{Definition}
\fi

\ifx\proposition\undefined
\newtheorem{proposition}[theorem]{Proposition}
\fi

\ifx\remark\undefined
\newtheorem{remark}{Remark}
\fi

\ifx\conjecture\undefined
\newtheorem{conjecture}[theorem]{Conjecture}
\fi

\ifx\factoid\undefined
\newtheorem{factoid}[theorem]{Fact}
\fi

\ifx\axiom\undefined
\newtheorem{axiom}[theorem]{Axiom}
\fi

\title{Repulsive Attention: \\ Rethinking Multi-head Attention as Bayesian Inference}

\author{Bang An$^{\dagger}$, Jie Lyu$^{\dagger}$, Zhenyi Wang$^{\dagger}$, Chunyuan Li$^{\mathsection}$ \\ \bf{Changwei Hu$^*$, Fei Tan$^*$, Ruiyi Zhang$^{\ddagger}$, Yifan Hu$^*$, Changyou Chen$^{\dagger}$} \\
  $^{\dagger}$State University of New York at Buffalo, $^{\mathsection}$Microsoft Research, Redmond \\
  $^*$Yahoo Research, $^{\ddagger}$Duke University \\
  \texttt{isbangan@gmail.com, ch237duke@gmail.com}\\
  \texttt{yifanhu@verizonmedia.com, changyou@buffalo.edu}
  }
\date{}

\begin{document}

\maketitle
\begin{abstract}
The neural attention mechanism plays an important role in many natural language processing applications. In particular, multi-head attention extends single-head attention by allowing a model to jointly attend information from different perspectives. 
However, without explicit constraining, multi-head attention may suffer from {\em attention collapse}, an issue that makes different heads extract similar attentive features, thus limiting the model's representation power. 
In this paper, for the first time, we provide a novel understanding of multi-head attention from a Bayesian perspective. 
Based on the recently developed particle-optimization sampling techniques, we propose a non-parametric approach that explicitly improves the repulsiveness in multi-head attention and consequently strengthens model's expressiveness. Remarkably, our Bayesian interpretation provides theoretical inspirations on the not-well-understood questions: why and how one uses multi-head attention. 
Extensive experiments on various attention models and applications demonstrate that the proposed repulsive attention can improve the learned feature diversity, leading to more informative representations with consistent performance improvement on multiple tasks.
\end{abstract}

\section{Introduction}
Multi-head attention~\cite{attention} is an effective module in deep neural networks, with impressive performance gains in many natural-language-processing (NLP) tasks. 
By extending a single head to multiple paralleled attention heads, the architecture is widely adopted to capture different attentive information and strengthen the expressive power of a model. 
\citet{zhouhan} applied the idea of multi-heads on self-attention and extract a 2-D matrix instead of a vector to represent different contexts of a sentence. The Transformer \cite{attention} and its variants such as BERT \cite{bert} are influential architectures solely based on multi-head attention, achieving state-of-the-art performance on plenty of NLP tasks. 
The key of multi-head attention is its ability to jointly attend to information from different representation subspaces at different positions, which results in multiple latent features depicting the input data from different perspectives.
However, there are no explicit mechanisms guaranteeing this desired property, leading to potential attention redundancy or attention collapse, which has been observed in previous research \cite{prune,kovaleva2019revealing}. 
Although there exist works by directly adding regularization on loss functions to encourage diversity in multi-head attention \cite{zhouhan, disagree}, the underlying working principle has not been well-validated, and performance improvement is limited. Furthermore, an important problem on why and how multi-head attention improves over its single-head counterpart is poorly understood.


In this paper, we provide a novel understanding of multi-head attention from a Bayesian perspective by adapting the deterministic attention to a stochastic setting. 
The standard multi-head attention can be understood as a special case of our framework, where attention-parameter updates between heads are independent, instead of sharing a common prior distribution. 
Based on our framework, attention repulsiveness could then be imposed by performing Bayesian inference on attention parameters with the recently developed particle-optimization sampling methods \cite{svgd}, which has been shown to be effective in avoiding mode collapse. 
These methods treat each head as a particle/sample, which is then optimized to approximate a posterior distribution of an attention model. 
With it, multiple heads are enforced to move to modes in the parameter space to be far from each other, thus improving the repulsiveness in multi-head attention and enhancing model's expressiveness. 
Our Bayesian interpretation also provides a theoretical understanding of the reason and benefits of applying multi-head attention. 
Experiments on various attention models demonstrate the effectiveness of our framework.

Our contributions are summarized as follow:
\begin{itemize}
    \item We provide a new understanding of multi-head attention from a Bayesian perspective, yielding a more principled and flexible interpretation of multi-head attention.
    \item Based on the recently developed particle-optimization sampling techniques, we propose an algorithm to explicitly encourage repulsiveness in multi-head attention without introducing extra trainable parameters or explicit regularizers. The proposed method can be implemented with an efficient end-to-end training scheme.
    \item Our Bayesian interpretation provides a theoretical foundation to understand the benefits of multi-head attention, which reveals the existence of an optimal number of attention heads in a specific model.
    \item We apply our approach on four attention models with a wide range of tasks. Experimental results show that repulsive attention improves the expressiveness of models, and yields consistent performance gains on all the tasks considered.
\end{itemize}

\section{Preliminaries}
\subsection{Multi-head Attention}
The attention mechanism aims at modeling dependencies among elements of a learned representation at different positions.
The two commonly used attention functions are additive attention \cite{zhouhan, align} and dot-product attention \cite{attention}. 
We review the popularly used dot-product attention below and defer the additive attention to Appendix \ref{app:add}.

\paragraph{Dot-product Attention} 
The multi-head scaled dot-product attention is used in the Transformer model \cite{attention}. The attention function for a single head is formulated as mapping a query and a set of key-value pairs to output as
\begin{align}
    \Amat_i =& \softmax(\Qmat_i \Kmat_i^T/\sqrt{d_k}), \Zmat_i = \Amat_i \Vmat_i \label{dot_single}\\
    \text{where } & \Qmat_i = \Qmat \Wmat_i^Q, \Kmat_i = \Kmat \Wmat_i^K, \Vmat_i = \Vmat\Wmat_i^V \notag
\end{align}
$\Qmat, \Kmat, \Vmat$ are matrices depicting the hidden representation of every word in one sentence ({\it i.e.} self-attention) or two sentences ({\it i.e.} inter-attention); $d_k$ is the dimension of key and query; $\Zmat_i$ is the attention feature of the input sentence from the $i$-th head; $\{\Wmat_i^Q, \Wmat_i^K,\Wmat_i^V\}$ are the corresponding learnable attention parameters. The $M$-head attention projects the queries, keys and values into $M$ subspaces with different learnable
linear projections. These attention functions are performed in parallel and are concatenated at last, resulting in a final latent representation:
\begin{align}\label{mul}
    &\multihead(\Qmat, \Kmat, \Vmat) = \Zmat \Wmat^O, \text{ with} \\
    &\Zmat = \concat(\Zmat_1, \cdots, \Zmat_M) \notag
\end{align}

\subsection{Particle-optimization Sampling}
Particle-optimization sampling is a recently developed Bayesian sampling technique that interactively transports a set of particles/samples to a target distribution $p$ by minimizing the KL divergence between the particle density and the target $p$. In our case, $p$ would be a posterior distribution, $p(\thetav | \mathcal{D}) \propto \exp(-U(\thetav))$, of the parameter $\thetav \in \mathbb{R}^d$, defined over an observed dataset $\mathcal{D} = \{D_k\}_{k=1}^N$. Here $U(\thetav)\triangleq -\log p(\mathcal{D} | \thetav) - \log p_0(\thetav) = - \sum_{k=1}^N\log p(D_k | \thetav)-\log p_0(\thetav)$ is called the potential energy with $p_0$ a prior over $\thetav$. In our case, the model parameter $\thetav$ could be one or several of the attention parameters such as $\Wmat_i^Q$. For simplicity, we will stick to $\thetav$ in the presentation. In particle-optimization sampling, a total of $M$ particles $\{\thetav^{(i)}\}_{i=1}^M$ are updated iteratively to approximate $p(\thetav | \mathcal{D})$. 
In this paper, we use two representative algorithms, the Stein Variational Gradient Descent (SVGD) and the Stochastic Particle-Optimization Sampling (SPOS), for sampling. 

\paragraph{SVGD} 
In SVGD \cite{svgd}, 
the $i$-th particle in the $(\ell+1)$-th iteration is updated with stepsize $\epsilon_{\ell+1}$ as
\begin{align}
    &\thetav_{\ell + 1}^{(i)} = \thetav_{\ell}^{(i)} + \epsilon_{\ell+1} \phi(\thetav_{\ell}^{(i)})\label{svgd1}\\ 
    \phi(\thetav_{\ell}^{(i)}) = &\frac{1}{M} \sum_{j=1}^M[-\kappa(\thetav_{\ell}^{(j)}, \thetav_{\ell}^{(i)})\nabla_{\thetav_{\ell}^{(j)}}U(\thetav_{\ell}^{(j)}) \notag \\ 
    &+ \nabla_{\thetav_{\ell}^{(j)}}\kappa(\thetav_{\ell}^{(j)}, \thetav_{\ell}^{(i)})]\label{svgd2}
\end{align}
where $\kappa(\cdot, \cdot)$ is a positive definite kernel ({\it e.g.}, RBF kernel). 
The two terms in $\phi$ play different roles: the first term drives the particles towards high density regions of $p(\thetav | \mathcal{D})$; whereas 
the second term acts as a repulsive force that prevents all the particles from collapsing together into local modes of $p(\thetav | \mathcal{D})$. 

\paragraph{SPOS} 
Though obtaining significant empirical success, under certain conditions, SVGD
experiences a theoretical pitfall, where particles tend to collapse. To overcome this, 
\citet{zhang2018stochastic} generalize SVGD to a stochastic setting by injecting random noise into particle updates. The update rule for particles $\thetav_{\ell}^{(i)}$ is 
{\fontsize{9}{10}\begin{align}\label{spos1}
    &\phi(\thetav_{\ell}^{(i)}) = \frac{1}{M} \sum_{j=1}^M[-\kappa(\thetav_{\ell}^{(j)}, \thetav_{\ell}^{(i)})\nabla_{\thetav_{\ell}^{(j)}}U(\thetav_{\ell}^{(j)}) +  \\ 
    & \nabla_{\thetav_{\ell}^{(j)}}\kappa(\thetav_{\ell}^{(j)}, \thetav_{\ell}^{(i)})] 
     -\beta^{-1}\nabla_{\thetav_{\ell}^{(i)}}U(\thetav_{\ell}^{(i)})+\sqrt{2\beta^{-1}\epsilon_{\ell}^{-1}} \xiv_{\ell}^{(i)} \notag
\end{align}}
where $\beta > 0$ is a hyperparameter, $\xiv_{\ell}^{(i)} \sim \mathcal{N}(\bm{0}, \bm{I})$ is the injected random Gaussian noise to enhance the ability of escaping local modes, leading to better ergodic properties compared to standard SVGD. 

\section{A Bayesian Inference Perspective of Multi-head Attention} \label{app:attention}

In this section, we interpret multi-head attention as Bayesian inference of latent representation via particle-optimization sampling. We denote $\bm{x}$ and $\bm{z}$ as the input and output (latent representation) of the attention model, respectively. The single-head attention can be written as a deterministic mapping $\bm{z} = f_{att}(\bm{x};  \thetav)$, with $\thetav$ the parameter of the mapping. Standard multi-head attention defines multiple parallel attention mappings, each endowed with independent parameters. The attention features are finally aggregated via a function $g(\cdot)$ as
\begin{align}
    \bm{z} = g(\bm{z}_1,...,\bm{z}_M), \quad \bm{z}_i =f_{att}(\bm{x}; \thetav_i)~. \label{det}
\end{align}
Next, we generalize \eqref{det} as a Bayesian inference problem for the latent representation $\bm{z}$.

\paragraph{Attention as Bayesian Inference} 
We first generalize the deterministic transformation, $\bm{z} = f_{att}(\bm{x}; \thetav)$, to a stochastic generative process as: 
\begin{align*}
    \thetav \sim p(\thetav|\mathcal{D}),~~\bm{z} = f_{att}(\bm{x}; \thetav)~,
\end{align*}
where a sample of the posterior of the global attention parameter $\thetav$, $p(\thetav|\mathcal{D}) \propto p(\mathcal{D} | \thetav)p(\thetav)$, is used as the parameter when generating the latent attention feature $\bm{z}$. 
Bayesian inference for attention then computes the predictive distribution $p(\bm{z} | \bm{x}, \mathcal{D})$ of the attentive latent representation  $\bm{z}$ for a new input $\bm{x}$ given the training data $\mathcal{D}$ by
    $p(\bm{z} | \bm{x}, \mathcal{D}) = \int \delta_{f_{att}(\bm{x}; \thetav)}(\bm{z}) p(\thetav | \mathcal{D})d\thetav~ \label{stoc}$, where $\delta_{\bm{z}}(\cdot)$ is the delta function with point mass at $\bm{z}$. 
%
To enable efficient evaluation of the integral, we adopt Bayesian sampling for approximation, {\it i.e.}, $p(\bm{z} | \bm{x}, \mathcal{D})$ is approximated by a set of $M$ samples/particles initialized from $p(\thetav|\mathcal{D})$, leading to the following generative process:
\begin{align}
    &\bm{z} = g(\bm{z}_1,...,\bm{z}_M) \label{re_mul}\\ 
    \bm{z}_i =f_{att}(&\bm{x}; \thetav_i), \quad \text{with} \quad \thetav_i\sim p(\thetav | \mathcal{D}) \notag
\end{align}
The above formulation defines a principled version of multi-head attention from a Bayesian view.
One can see that \eqref{re_mul} reduces to the standard multi-head attention if all $\thetav_i$ are treated independently without sharing the common parameter distribution $p(\thetav|\mathcal{D})$. 
In other words, our reformulation of multi-head attention is a stochastic generative process, thus is more general. Furthermore, efficient end-to-end learning can be performed by conducting repulsive Bayesian sampling for all parameters $\{\thetav_i\}_{i=1}^M$, which consequently could diversify the attention features $\{\bm{z}_i\}_{i=1}^M$.

\section{Repulsive Attention Optimization}
The Bayesian multi-head attention in \eqref{re_mul} further inspires us to develop the repulsive attention.   
The idea is to learn to generate repulsive samples from the posterior $p(\thetav|\mathcal{D})$. 
We propose to adopt the particle-optimization sampling methods, which could explicitly encourage repulsiveness between samples.
In our algorithm, the parameter of $p(\bm{z} | \bm{x}; \thetav)$ for each head is considered as one particle. 
Following the particle-optimization rules, $M$ heads $\{\thetav_i\}_{i=1}^M$ are updated iteratively to approximate the posterior distribution of attention parameter $p(\thetav | \mathcal{D})$.  

\subsection{Learning Repulsive Multi-head Attention}
We propose to learning repulsive attention by replacing the standard updates of attention parameters via stochastic gradient descent (SGD) with particle-optimization sampling methods while keeping the multi-head attention model unchanged. This procedure forms an efficient end-to-end training scheme similar to standard attention learning. 
To be specific, in standard multi-head attention, the parameter of every head is updated independently according to the respective gradient of a loss function. To achieve repulsive multi-head attention, we follow the particle-optimization sampling update rule ({\it e.g.} \eqref{svgd1} and \eqref{svgd2}) to update the parameter of every head while keeping the update for the remaining parameters via SGD unchanged. Equations \eqref{svgd2} and \eqref{spos1} can be viewed as modified gradients with explicit repulsive force and can be integrated into any optimizer, {\it e.g.}, Adam \cite{adam}. Note that $\nabla_{\thetav_{\ell}^{(i)}}U(\thetav_{\ell}^{(i)})$ equals to the gradient of $\thetav_{\ell}^{(i)}$ in standard multi-head attention when the negative log-likelihood is used as the loss function and the prior of $\thetav^{(i)}$ is assumed to be uniform. The learning algorithm is illustrated in Algorithm \ref{alg:1}. In practice, the update of $M$ heads can be performed in parallel with efficient matrix operations.

\begin{algorithm} 
\caption{Repulsive Multi-head Attention}
\label{alg:1}
{\bf Input:} Initialized $M$-head attention model $\mathcal{A}$ with attention parameters $\bm{\Theta}_0 = \{\thetav_i\}_{i=1}^M$ and all the other parameters $\bm{\Omega}_0$; Training data $\mathcal{D} = \{D_k\}_{k=1}^N = \{(\xv_k, y_k)\}_{k=1}^N$; 

{\bf Output:} Optimized attention model with learned parameters $\hat{\bm{\Theta}}$ and $\hat{\bm{\Omega}}$;

{\bf Train:}
\begin{algorithmic}[l]
\FOR{iteration $\ell$}
 \STATE {\bf forward}: $\hat{y}_k = \mathcal{A}(\xv_k; \bm{\Theta}_{\ell},\bm{\Omega}_{\ell}), \forall k$;
 \STATE calculate loss: $\mathcal{L}(\{\hat{y}_k\}, \{y_k\})$;
 
\STATE {\bf backward} and calculate gradients:
\STATE gradient of $\bm{\Omega}_{\ell}$: $\varphi(\bm{\Omega}_{\ell}) \gets \nabla_{\bm{\Omega}_{\ell}} \mathcal{L}$ 
 \FOR{attention head $i$}
    \STATE calculate $\phi(\thetav_{\ell}^{(i)})$ with Eq (\ref{svgd2}) or (\ref{spos1});
    \STATE gradient of $\thetav_{\ell}^{(i)}$:$\varphi(\thetav_{\ell}^{(i)})\gets \epsilon_{\ell} \phi(\thetav_{\ell}^{(i)})$;
 \ENDFOR
 \STATE update parameters:
 \STATE $\bm{\Omega}_{\ell+1} \gets \text{Optimizer}(\bm{\Omega}_{\ell}, \varphi(\bm{\Omega}_{\ell}))$
 \STATE $\bm{\Theta}_{\ell+1} \gets \text{Optimizer}(\bm{\Theta}_{\ell}, \varphi(\bm{\Theta}_{\ell}))$
\ENDFOR

\end{algorithmic}
\end{algorithm}

\subsection{In-depth Analysis}\label{sec:why}
\paragraph{Why Multi-head Attention?}
Our Bayesian interpretation of the attention mechanism naturally provides an answer to the question of why one needs multi-head attention. By treating each head as one sample, adopting multiple heads means using more samples to approximate an underlying posterior distribution. The question comes to should one use more heads (samples). Intuitively this seems to be true because more samples typically make more accurate approximations. However, this could not be the case in practice. The reason might be two-fold: $\RN{1})$ {\em{Overfitting}}: Learning with a limited amount of data could easily causes overfitting, thus requiring a smaller model (less attention heads); $\RN{2})$ {\em{Numerical error}}: Our proposed method to update samples (attention-head parameters) is essentially a discrete numerical method of the corresponding continuous-time partial differential equation, {\it i.e.}, the samples are not exact samples from the target distribution. Thanks to the recently developed theory for particle-optimization sampling \cite{zhang2018stochastic}, one can conclude that more heads could accumulate more numerical errors, leading to performance deterioration. More formally, when using particles to approximate a target posterior distribution, there exists a gap (approximation error) between the particle distribution and the true distribution \cite{zhang2018stochastic}. This approximation error, when applied to our setting, approximately scales in the order of $O(\frac{1}{\sqrt{M}} + M\epsilon_0^{1/2} + e^{-\sum_{\ell}\epsilon_{\ell}})$. Please refer to Theorem 10 in \cite{zhang2018stochastic} for a formal description.

\paragraph{How Many Heads are Enough?}
The above error bound suggests that there is a trade-off between approximation accuracy and the number of heads $M$. Specifically, we have $\RN{1})$ when $M$ is small, the term $\frac{1}{\sqrt{M}}$ in the bound would dominate, leading to decreasing errors (increasing performance) with increasing $M$; $\RN{2})$ when $M$ is large enough, the term $M\epsilon_0^{1/2}$ dominates, suggesting that larger $M$ could actually increase the approximation error (decreased performance). These phenomena are consistent with our experimental results. We note that an exact form of the optimal $M$ is not available due to a number of unknown constants (omitted in the big-O notation). Therefore, one should seek other ways such as cross-validation to choose a good $M$ in practice. Our argument also aligns with recent research, which found that more heads do not necessarily lead to better performance \cite{sixteen}.

\section{Experiments}
We demonstrate the effectiveness of our method with representative multi-head attention models on a broad range of tasks including sentence classification, machine translation, language modeling and text generation. This section summarizes key results on different models. 
More detailed experiment settings and analysis are deferred to the appendix.
To apply our approach, only the learning method of multi-head attention is adapted.

\begin{figure*}[t]
    \centering
    \subfloat[detailed attentions]{{\includegraphics[width=0.46\textwidth]{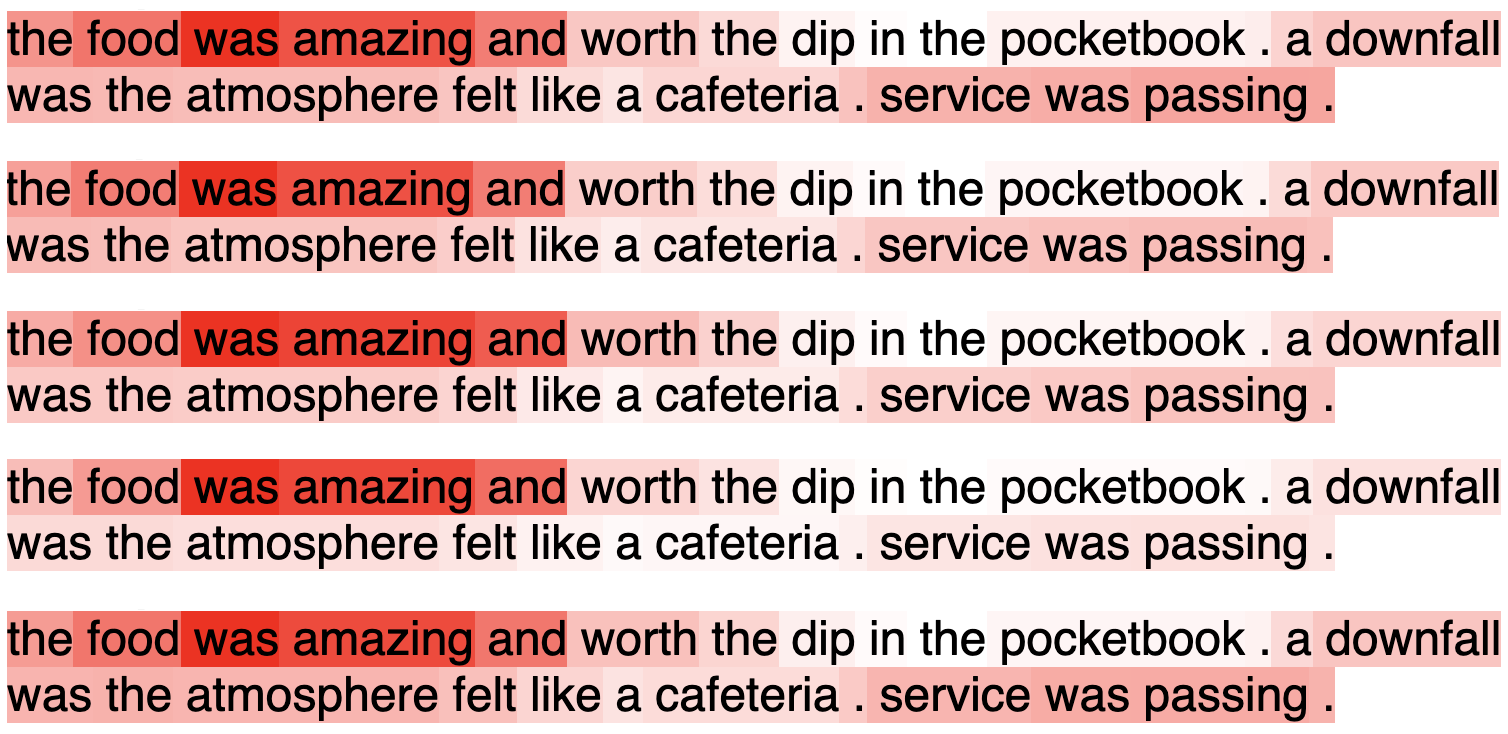}}} 
    \qquad
    \subfloat[detailed attentions]{{\includegraphics[width=0.46\textwidth]{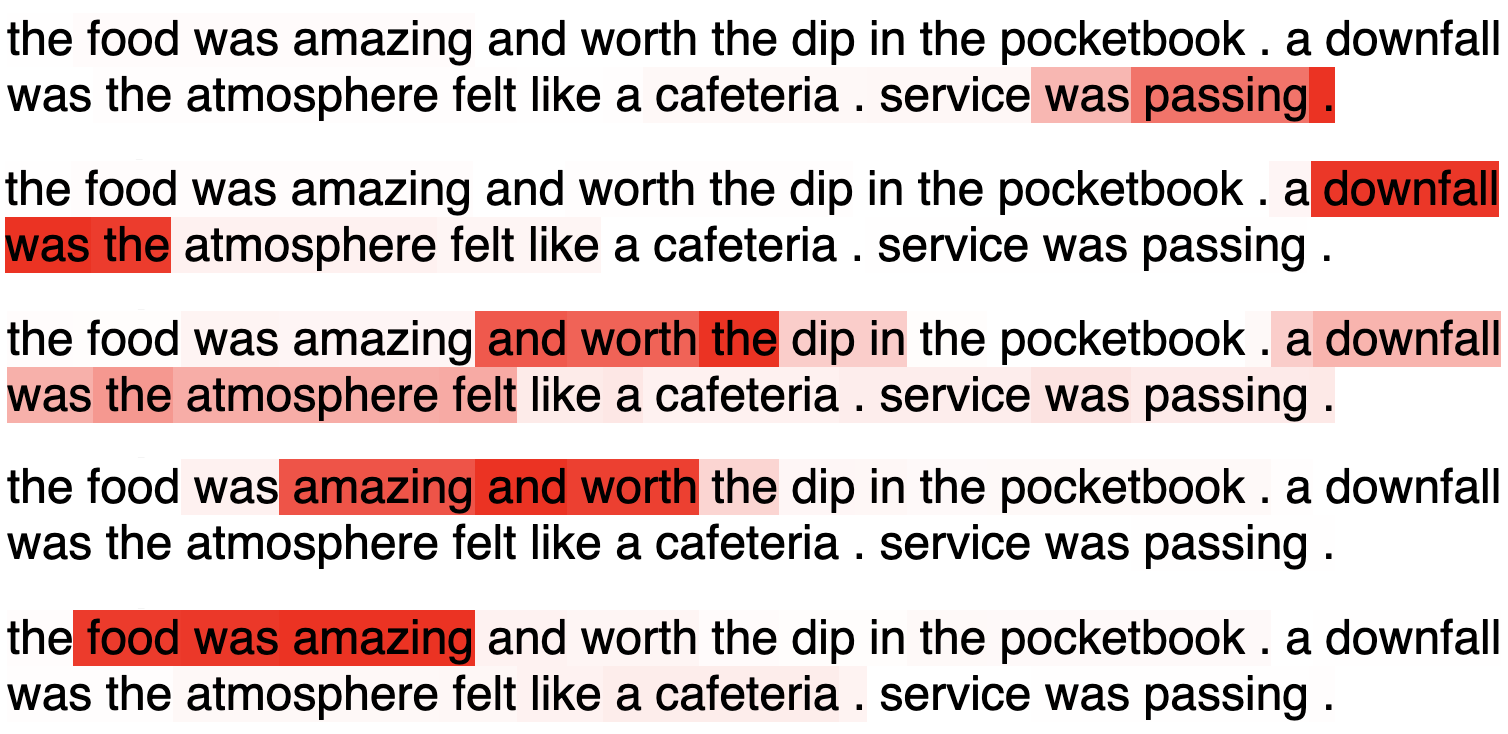}}} \\
    \subfloat[standard multi-head attention]{{\includegraphics[width=0.46\textwidth]{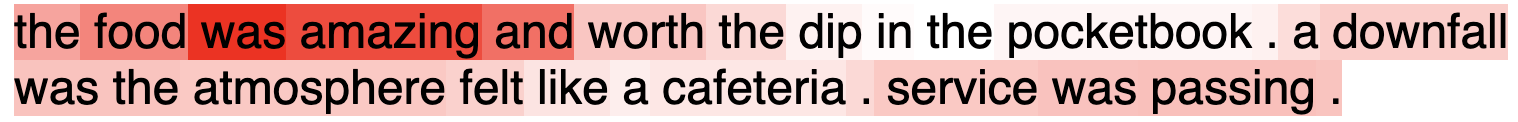}}} 
    \qquad
    \subfloat[repulsive multi-head attention]{{\includegraphics[width=0.46\textwidth]{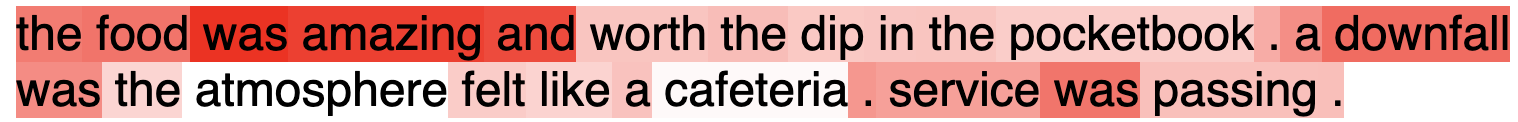}}} 
    \caption{Attention heatmaps of a 4-star Yelp review. Results on the left is from the standard multi-head attention, and the result on the right is from our repulsive multi-head attention. (a) and (b) shows detailed attention maps taken by 5 out of 30 rows of the matrix embedding, while (c) and (d) shows the overall attention by summing up all 30 attention weight vectors.}
    \label{fig:vis}
    \vspace{-0.3cm}
\end{figure*}
\subsection{Self-attentive Sentence Classification} \label{sec:classification}

\begin{table}[]
\small
    \centering
    \begin{tabular}{l c c}
    \toprule
    \textbf{Models} & \textbf{Acc(\%) }& \textbf{Dist}\\
    \midrule
    \multicolumn{3}{c}{\quad \quad  Age}\\
    \midrule
    BiLSTM + MA & 81.47  & 0.129 \\
    BiLSTM + MA + R & 81.30 & 0.178 \\
    BiLSTM + RMA (SVGD) & 81.82 & 0.492 \\
    BiLSTM + RMA (SPOS) & \textbf{82.55} & 0.461 \\
    \midrule
    \multicolumn{3}{c}{\quad \quad  Yelp}\\
    \midrule
    BiLSTM + MA & 69.3  & 0.246 \\
    BiLSTM + MA + R & 70.2 & 0.536 \\
    BiLSTM + RMA (SVGD) & 71.2 & 1.602 \\
    BiLSTM + RMA (SPOS) &\textbf{ 71.7} & 1.655 \\
    \midrule
    \multicolumn{3}{c}{\quad \quad  SNLI}\\
    \midrule
    BiLSTM + MA & 83.79  & 1.293 \\
    BiLSTM + MA + R & 84.55 & 1.606 \\
    BiLSTM + RMA (SVGD) & 84.58 & 1.688 \\
    BiLSTM + RMA (SPOS) & \textbf{84.76} & 1.370 \\

    \bottomrule
    \end{tabular}
    \caption{Performance (accuracy) comparison on Age, Yelp and SNLI dataset. Dist: the average 2-norm distance between each pair of the latent representation encoded from different heads on test set. MA: standard multi-head attention. RMA: proposed repulsive multi-head attention. R: regularization approach.}
    \label{tab:self-att sentence}
\end{table}


\paragraph{Model \& Baselines}
We first apply our method to the self-attentive sentence classification model \cite{zhouhan} which combines BiLSTM with additive attention to learn the sentence embedding and then does classification on it. We compare our method with the one using the standard multi-head attention (BiLSTM + MA) and the one applying the Frobenius regularization (BiLSTM + MA + R) on it to introduce diversity as in \citet{zhouhan}.

\paragraph{Tasks \& Datasets} 
Following \citet{zhouhan}, three sentence classification tasks including author profiling, sentiment analysis, and textual entailment are evaluated on the Age, Yelp, and SNLI datasets respectively. 



\paragraph{Results}
As shown in Table \ref{tab:self-att sentence}, with the proposed repulsive multi-head attention, the model achieves higher accuracy on all three tasks. Especially on the sentiment analysis task which often contains multiple aspects in one sentence. Our methods also outperform the regularization method proposed in \citet{zhouhan}. With different particle-optimization rules, SPOS is able to achieve better performance due to its extra advance discussed by \citet{zhang2018stochastic}. 
We further evaluate the diversity of multiple heads by calculating the average distance between each pair of latent representations. 
Results show that our methods indeed enforce heads to be more diverse, compared with the standard multi-head attention. The less diverse of the regularization-based method also indicates the validness of our argument in Appendix~\ref{app:regu}. 
 
\paragraph{Repulsive-attention visualization} 
We further visualize attention maps in the learned sentence embedding space in Figure \ref{fig:vis}. 
It is interesting to see attention collapse indeed happens in the standard multi-head attention, where almost all heads focus on one single factor \textit{"amazing"}. On the contrary, the proposed method is able to capture multiple key factors in the review that are strong indicators of the sentiment behind the sentence. For example, \textit{"downfall"} and \textit{"service was passing"} are key factors for this 4-star review captured by our repulsive multi-head attention, which are missed by the standard attention. 
The full attention heatmaps of all 30 heads and more examples are in Appendix \ref{app:vis}.

\subsection{Transformer-based Neural Translation}\label{sec:transformer}
\paragraph{Model \& Baselines}
The Transformer \cite{attention} is a representative multi-head attention based model. We apply the proposed repulsive multi-head attention (RMA) on it and compare our method with the original one (MA) and the disagreement regularization method (R) \cite{disagree} which encourages the diversity in attention by a cosine similarity penalty on attention outputs.

\paragraph{Tasks \& Datasets}
Following \citet{attention}
, we apply Transformer for machine translation, with two standard translation datasets: the IWSLT14 German-to-English (De-En) dataset , and the WMT14 English-to-German (En-De) dataset. 

\begin{table}[t]
\small
    \centering
    \begin{tabular}{l c c}
        \toprule
        \textbf{Models} & \textbf{BLEU} & \textbf{Time}\\
        \midrule
        \multicolumn{3}{c}{\quad \quad  IWSLT14 De-En}\\
        \midrule
        Transformer-\texttt{small}-MA & 34.4 & 1 \\
        Transformer-\texttt{small}-MA + R & 34.9 & 1.29 \\
        Transformer-\texttt{small}-RMA & \textbf{35.2} & 1.13 \\
        \midrule
        \multicolumn{3}{c}{\quad \quad  WMT14 En-De}\\
        \midrule
        Transformer-\texttt{base}-MA & 27.3 & 1\\
        Transformer-\texttt{base}-MA + R & 28.2 & 1.35 \\
        Transformer-\texttt{big}-MA &  \textbf{28.4} & - \\
        Transformer-\texttt{base}-RMA & \textbf{28.4} & 1.18\\
        \bottomrule
    \end{tabular}
    \caption{Translation Performance on IWSLT14 De-En and WMT14 En-De Datasets. MA: standard multi-head attention. RMA: proposed repulsive multi-head attention. R: regularization approach. Time: relative training time of every step versus MA.}
    \label{tab:transformer_main}
    \vspace{-0.3cm}
\end{table}
\paragraph{Results}
Results are presented in Table \ref{tab:transformer_main}. With the repulsive multi-head attention, Transformer models achieve noticeable improvement on the BLEU score on both datasets, compared with both baselines.
It is also encouraging to see that the Transformer-\texttt{base}-RMA with a much smaller model achieves comparable performance as Transformer-\texttt{big}.
As for training time, our approach takes slightly more time than the baseline, but is much more efficient than the regularization approach. 

\paragraph{Which attention module to be diversified?}
We conduct extra experiments on Transformer-\texttt{small} to investigate which attention module benefits most from the repulsiveness. Results (see Appendix \ref{app:Transformer-abl}) suggest that diversifying different attention module benefits differently. Remarkably, only diversifying the attention in the first layer is able to achieve comparable performance to the case of diversifying attention in all layers, with little computational time increased. 
This finding suggests that the repulsiveness in the first layer's attention plays an important role for modelling language. 

\paragraph{Redundancy in heads}
\begin{figure}
    \centering
    \subfloat[Transformer-\texttt{base}-MA]{{\includegraphics[width=0.5\linewidth]{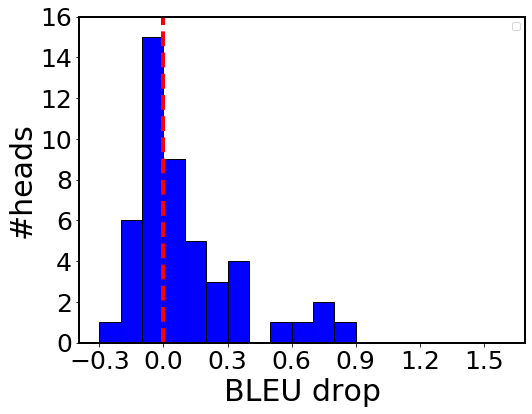}}} 
    \subfloat[Transformer-\texttt{base}-RMA]{{\includegraphics[width=0.5\linewidth]{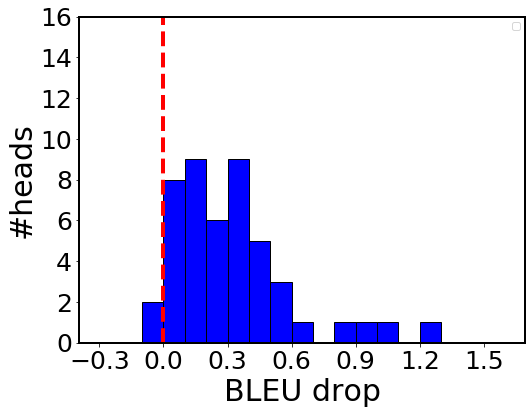}}} 
    \caption{Distribution of heads by performance drop after masking at test time. The redundancy of heads in RMA is much less.}
    \label{fig:redundancy}
    \vspace{-0.3cm}
\end{figure}
The redundancy problem in attention has been observed in recent works \cite{sixteen}, that a large
percentage of attention heads can be removed at test time without significantly
impacting performance. 
Following \citet{sixteen}, we analysis the redundancy in Transformer by ablating each head at testing and evaluating the performance drop. The more drops, the more important of the head. 
Figure \ref{fig:redundancy} shows that the majority of heads in standard multi-head attention are redundant for the performance is comparable before and after masking. 
However, the repulsive attention largely alleviates the redundancy. 
More interestingly, there are a lot of counter-intuitive cases in standard attention: removing a head results in an increase in performance. However, this does not seem to happen in repulsive attention model, indicating better leveraging of the superior expressiveness of multi-head mechanism.

\subsection{Language Representation Learning} \label{sec:electra}
\begin{table*}[t]
\small
    \centering
    \setlength\tabcolsep{4pt}
    \begin{tabular}{c | l c c c c c c c c @{\hspace{1.5\tabcolsep}} c @{\hspace{1.7\tabcolsep}} c}
    \toprule \small
    \textbf{Pre-training Data}     & \textbf{Model} & \textbf{Prams} & \textbf{CoLA} & \textbf{SST} & \textbf{MRPC} & \textbf{STS} & \textbf{QQP} & \textbf{MNLI} & \textbf{QNLI} & \textbf{RTE} & \textbf{Avg.} \\
    \midrule
        \multirow{5}{*}{\shortstack{Wikipedia \\ + \\BooksCorpus}} & TinyBERT & 14.5M & 51.1 & 93.1 & 82.6 & 83.7 & 89.1 & 84.6 & 90.4 & 70.0 & 80.6\\ 
         & MobileBERT & 25.3M & 51.1 & 92.6 & 84.5 & 84.8 & 88.3 & 84.3 & 91.6 & 70.4 & 81.0\\
         & GPT & 117M & 45.4 & 91.3 & 75.7 & 80.0 & 88.5 & 82.1 & 88.1 & 56.0 & 75.9 \\
         & BERT-Base & 110M & 52.1 & 93.5 & 84.8 & 85.8 & 89.2 & 84.6 & 90.5 & 66.4 & 80.9 \\
         & ELECTRA & 14M & 54.6 & 89.1 & 83.7 & 80.3 & 88.0 & 79.7 & 87.7 & 60.8 & 78.0 \\
    \midrule
       \multirow{2}{*}{\shortstack{OpenWebText}} & ELECTRA & 14M & 56.2 & \textbf{88.3} & 87.5 & 86.8 & 88.1 & 78.6 & 87.4 &\textbf{67.3} / 71.4 & 80.0 / 80.5  \\
        & \quad + RMA  & 14M & \textbf{59.4} & 87.1 & \textbf{87.9} &\textbf{87.0} & \textbf{88.6} &\textbf{79.3} & \textbf{87.8} & 64.9 / \textbf{73.1} & \textbf{80.3} / \textbf{81.3} \\
    \bottomrule
    \end{tabular}
    \caption{Results on the GLUE test set. For RMA, repulsive attention is only applied to pre-training. For RTE, the left value is fine-tuned from pre-trained models, the right value is from intermediate task training.}
    \label{tab:electra}
    \vspace{-0.3cm}
\end{table*}

\paragraph{Model}
ELECTRA \cite{electra} is an efficient approach to self-supervised language representation learning. 
It consists of two networks, {\em Generator} and {\em Discriminator}, both of which are parameterized by Transformers. The pre-trained Discriminator is used in various downstream tasks via fine-tuning.
We apply the proposed repulsive multi-head attention to ELECTRA (small setting) in the pre-training stage.
We only make the first layer attention of Discriminator to be repulsive, according to the finding in Section~\ref{sec:transformer} that the diversity in the first attention layer of Transformer benefits the most. 
\paragraph{Tasks \& Dataset}
We train ELECTRA models on OpenWebText Corpus due to the data used in \citet{electra} is not publicly available.
The pretrained models are then fine-tuned and evaluated on the General Language Understanding Evaluation (GLUE) \cite{GLUE} benchmark on eight datasets \cite{cola, sst, mrpc, sts, mnli, qnli, rte1,rte2,rte3,rte4}.

\paragraph{Results}
Results are shown in Table \ref{tab:electra}.  For each task, we perform single-task fine-tuning 50 times, and report the averaged results. The training time with and without repulsive attention is almost the same.
It shows that repulsive attention improves the baseline results \cite{electra} in seven out of eight tasks on GLUE, and the gains are larger especially on MNLI (the largest dataset on GLUE) and CoLA . This suggests that repulsive attention can yield better language representations. 
Since MNLI and RTE are both entailment tasks, following~\citet{electra} and ~\citet{intermediate}, we use intermediate task training for RTE. We first fine-tune the pre-trained model on MNLI, then continuously fine-tune it on RTE. The repulsive attention outperforms the baseline method by a large margin in this setting.
This is probably because the repulsive attention particularly favor large data variability (e.g., MNLI dataset), where different aspects of data can be uniquely represented in different heads. 



\subsection{Graph-to-Text Generation}\label{sec:GW}
\paragraph{Model \& Baselines}
GraphWriter \cite{Graphwriter} is a knowledge-graph-to-text model, which aims at generating coherent multi-sentence abstract given a knowledge graph and a title. 
There is a Transformer-style encoder defined with graph attention modules \cite{GAT} that could also be easily adapted to our method. 
We compare our method with the original one that has the standard multi-head attention, and the one with the cosine similarity regularization on attention parameters in encoder layers.
\paragraph{Tasks \& Datasets \& Metrics}
Experiments are conducted on the Abstract GENeration DAtaset (AGENDA) \cite{Graphwriter}, a dataset of knowledge graphs paired with scientific abstracts. 
We evaluate the quality of abstracts with 3 major metrics: 
BLEU (uni-gram to 4-gram BLEU) \cite{BLEU}, METEOR \cite{METEOR}, ROUGE \cite{ROUGE}. In ROUGE, the unigram and bigram overlap (ROUGE-1 and ROUGE-2) are a proxy for assessing informativeness and the longest common
subsequence (ROUGE-L) represents fluency. 

\begin{table}[]
\small
    \centering
    \begin{tabular}{l c c c}
        \toprule
       \textbf{ Metrics} & \textbf{GW} & \textbf{+ R} & \textbf{+ RMA} \\
        \midrule
        BLEU-1 & 42.56 & 42.25 & \textbf{45.60} \\
        BLEU-2 & 27.64 & 27.98 & \textbf{29.96} \\
        BLEU-3 & 19.27 & 19.77 & \textbf{21.07} \\
        BLEU-4 & 13.75 & 14.21 & \textbf{15.12} \\
        METEOR & 18.11  & 18.61 & \textbf{19.52}\\
        ROUGE-1 & 35.80 & 37.24 & \textbf{38.23}\\
        ROUGE-2 & 16.83 & 17.78 & \textbf{18.39}\\
        ROUGE-L & 27.21 & 26.90 & \textbf{28.55}\\
        \bottomrule
    \end{tabular}
    \caption{Automatic evaluations of generation systems on test set of AGENDA.}
    \label{tab:GW}
\end{table}



\begin{table}[]
\small
    \centering
    \begin{tabular}{l c c c}
    \toprule
         & Win & Lose & Tie  \\
         \midrule
         Structure & 51\% & 12\% & 37\% \\
         Informativeness & 66\% & 13\% & 21\% \\
         Grammar & 37\% & 17\% & 46\% \\
         Overall & 65\% & 14\% & 21\% \\
    \bottomrule
    \end{tabular}
    \caption{Human judgments of GraphWriter with and without repulsive attention.}
    \label{tab:human}
    \vspace{-0.3cm}
\end{table}

\paragraph{Results}
The results are shown in Table \ref{tab:GW}. 
The GraphWriter model with repulsive multi-head attention significantly outperforms the original model and regularization approach in all metrics.  
Especially, the higher recall score in ROUGE shows that there are more N-grams across the reference abstracts that can be found in the generated abstracts.
Similar observations are noticed when analyzing the generated examples in detail (an example is illustrated in Appendix \ref{app:GW}). 
\citet{Graphwriter} pointed out one limitation of their model is 40\% of entities in the knowledge graphs do not appear in the generated text. 
With the repulsive attention, remarkably, the GraphWriter model is observed to perform much better with a 10\% improvement on the knowledge graph coverage and fewer repeat clauses. 

\paragraph{Human Evaluation}
To further illustrate the improvement of using diverse attention, we conduct human evaluation. 
Following~\citet{Graphwriter}, we give 50 test datapoints to experts (5 computer science students) and ask them to provide per-criterion judgments for the generated abstracts.
Comparisons of the two methods from 4 aspects are shown in Table \ref{tab:human}. 
The human judgment indicates that the repulsive attention improves both the structure and informativeness of generated abstracts significantly, which is consistent with the automatic evaluation and our observations.

\subsection{On the Number of Attention Heads}
\begin{figure}[t!]
    \centering
    \subfloat[]{{\includegraphics[width=0.5\linewidth]{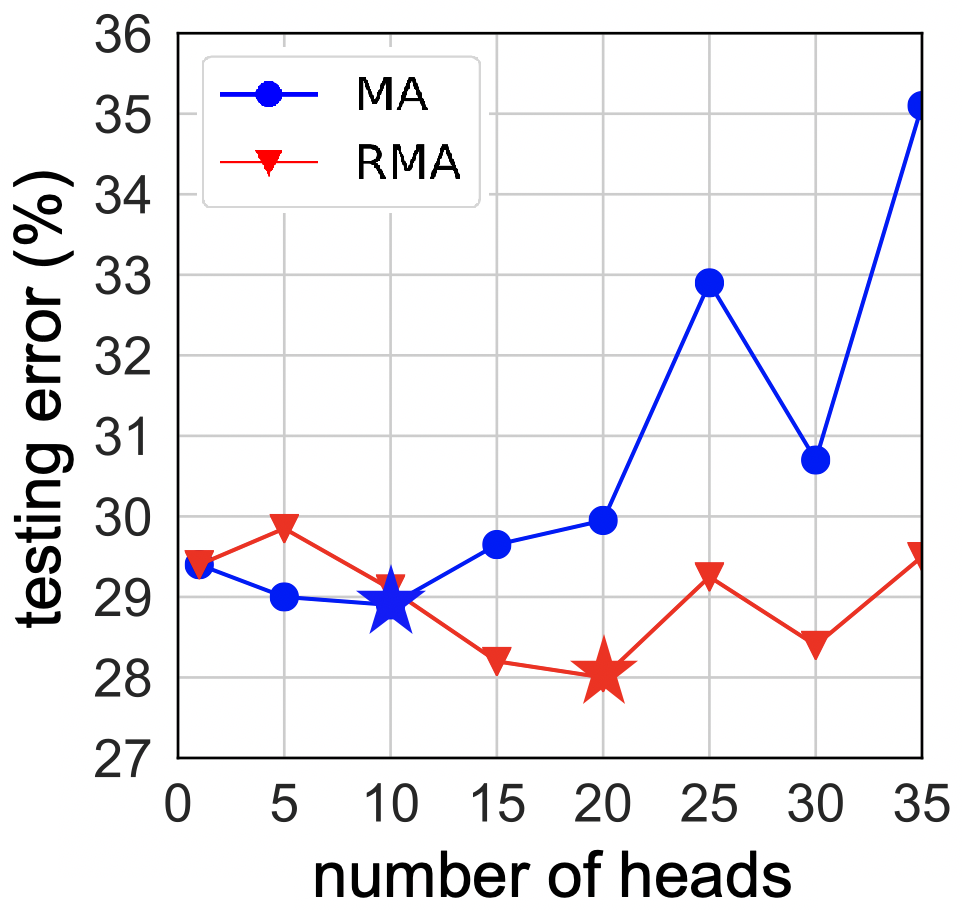}}} 
    \subfloat[]{{\includegraphics[width=0.5\linewidth]{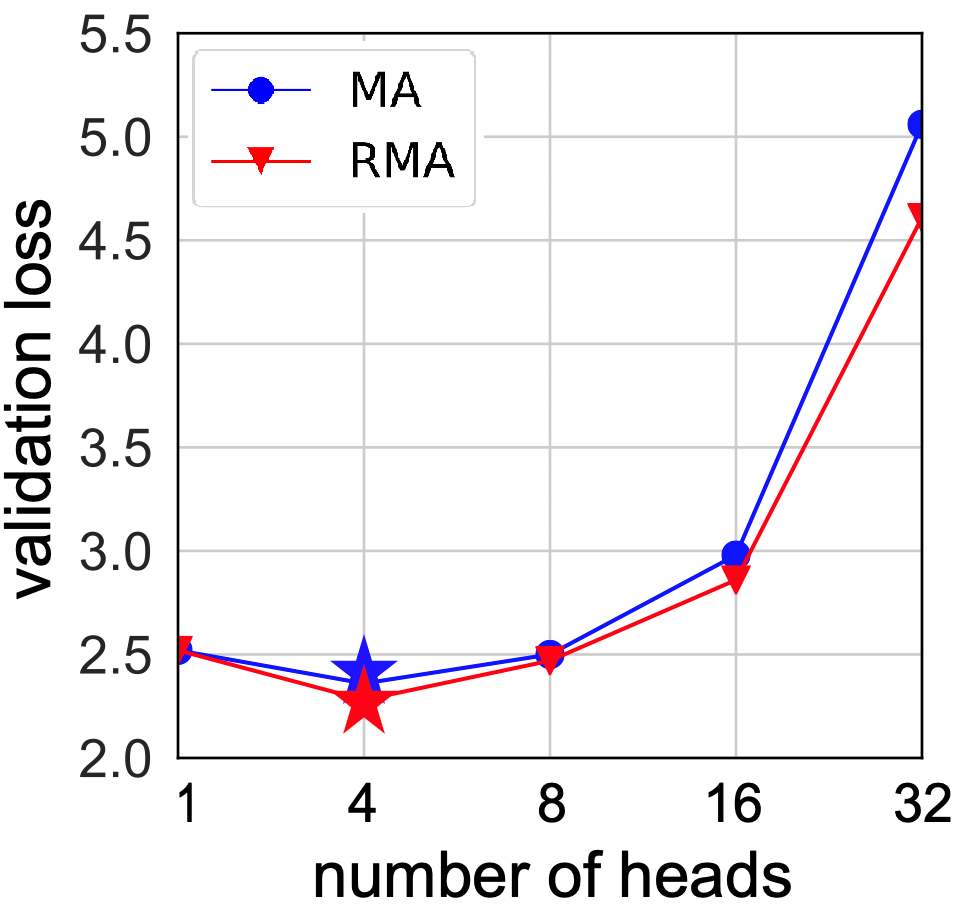}}} \vspace{-0.3cm}
    \caption{Demonstration on performance difference with different number of heads. (a) Testing error of the self-attentive sentence classification model on Yelp dataset. (b) Negative log likelihood loss of Transformer-\texttt{small} on IWSLT14 De-En dataset.  }
    \label{fig:head}
    \vspace{-0.3cm}
\end{figure}

Our analysis in Section~\ref{sec:why} suggests the existence of the optimal number of attention heads. To verify this, we further conduct experiments on sentence classification and translation tasks by varying the number of attention heads in models. The results are shown in Figure~\ref{fig:head}. 
The model error/loss first decreases then increases w.r.t.\! $M$, the number of attention heads. The optimal $M$ are around 20 and 4 for the sentiment analysis and the Transformer, respectively. Interestingly, the Transformer degrades quickly as the number of heads increases. This might because the constant corresponding to the $O(M\epsilon_0)$ term in the bound is too large, making this term quickly dominate with increasing $M$. Furthermore, it is also observed that the standard multi-head attention follows the same trend, but performs much worse and is more sensitive to the $M$. This indicates the benefit of Bayesian modeling, which could usually stabilize a model better.


\section{Related Work}
We provide a first explanation of multi-head attention from a Bayesian perspective, and propose particle-optimization sampling for repulsive attention. 
Most previous works aim at improving attention diversity with regularization-based methods, {\it e.g.}, the Frobenius regularization on attention weights in \citet{zhouhan} and the cosine similarity regularization on attention outputs in \citet{disagree}. These works focus on a particular model and the underlying working principle has not been well-validated. Our approach is a principled one that is more interpretable and widely applicable. 

The attention collapse belongs to a feature-overlapping problem, which also happens in other areas. 
Some works tackle this problem by changing architectures, for example ResNet \cite{resnet} and DenseNet \cite{densnet}
implicitly reduce feature correlations by summing
or concatenating activation from previous layers.
There are also works done by altering the training method as we do. 
\citet{DSD} adopt the dropout mechanism and propose a dense-sparse-dense training flow, for
regularizing deep neural networks. 
\citet{RePr} attempt addressing the unnecessary overlap in features captured by image filters with pruning-restoring scheme in training.
To our knowledge, we are the first to tackle the attention-feature overlap problem from a Bayesian view with a principled interpretation.

\section{Conclusion}
We propose a principled way of understanding multi-head attention from a Bayesian perspective. 
We apply particle-optimization sampling to train repulsive multi-head attention with no additional trainable parameters nor explicit regularizers. Our Bayesian framework explains the long-standing question of why and how multi-head attention affects model performance. Extensive experimental results on representative attention models demonstrate that our approach can significantly improve the diversity in multi-head attention, resulting in more expressiveness attention models with performance improvement on a wide range of tasks. 

\section*{Acknowledgements}
We sincerely thank all the reviewers for providing valuable feedback. This paper is supported by the Verizon Media FREP program.

\bibliographystyle{acl_natbib}
\bibliography{emnlp2020}

\appendix

\newpage
\twocolumn

\input{appendix}

\end{document}

%% file: mlVecMat.tex
\newcommand{\RN}[1]{%
	\textup{\lowercase\expandafter{\it \romannumeral#1}}%
}

\def\tanh{\textsf{tanh}} 
\def\softmax{\textsf{Softmax}} 
\def\multihead{\textsf{Multi-head}} 
\def\concat{\textsf{Concat}} 

\newcommand{\beq}{\vspace{0mm}\begin{equation}}
\newcommand{\eeq}{\vspace{0mm}\end{equation}}
\newcommand{\beqs}{\vspace{0mm}\begin{eqnarray}}
\newcommand{\eeqs}{\vspace{0mm}\end{eqnarray}}
\newcommand{\barr}{\begin{array}}
\newcommand{\earr}{\end{array}}

\newcommand{\Amat}[0]{{{\bf A}}}

\newcommand{\Hmat}{{\bf H}}

\newcommand{\Kmat}{{\bf K}}

\newcommand{\Qmat}{{\bf Q}}

\newcommand{\Vmat}[0]{{{\bf V}}}
\newcommand{\Wmat}[0]{{{\bf W}}}

\newcommand{\Zmat}{{\bf Z}}

\newcommand{\av}{{\boldsymbol{a}}}

\newcommand{\vv}{\boldsymbol{v}}

\newcommand{\xv}{\boldsymbol{x}}

\newcommand{\zv}{\boldsymbol{z}}

\newcommand{\thetav}{\boldsymbol{\theta}}

\newcommand{\xiv}[0]{{\boldsymbol{\xi}}}

\newcommand{\R}{\mathbb{R}}

\newtheorem{theorem}{Theorem} 
\newtheorem{lemma}{Lemma}
\newtheorem{proposition}[theorem]{Proposition}
\newtheorem{corollary}{Corollary}

\newenvironment{proof}[1][Proof]{\begin{trivlist}
\item[\hskip \labelsep {\bfseries #1}]}{\end{trivlist}}

%% file: appendix.tex
\section{Additive Attention}\label{app:add}
First proposed by \cite{align}, additive attention uses a one-hidden layer feed-forward network to calculate the attention alignment. We use the attention function in \citet{zhouhan}, which is also a self-attention, as an example. It aims at extracting the latent representation of a sentence. The single-head attention function is: 
\begin{align*}
    \av = \softmax(\vv^{\top} \tanh(\Wmat \Hmat^{\top})), \zv = \av \Hmat \label{add_single}
\end{align*}
where $\Hmat \in \R^{n\times d} $ is the hidden state matrix of a sentence with $n$ words, every word is embedded in a $d$ dimensional vector. $\vv \in \R^{1\times n} $ is the normalized alignment score vector for each word. $\Wmat \in \mathbb{R}^{d_{a}\times d} $ and $\bm{v}\in \mathbb{R}^{d_a\times 1} $ are attention parameters. The final sentence representation vector $\zv$ is a weighted sum of words' hidden states weighted by attention vector. In order to capture overall semantics of the sentence instead of a specific component, multi-head attention could be applied as
\begin{equation*}
    \Amat = \softmax(\Vmat^{\top} \tanh(\Wmat \Hmat^{\top})) \label{add}, \Zmat = \Amat \Hmat
\end{equation*}
where $\Vmat \in \mathbb{R}^{d_{a}\times M}$ is the matrix performs $M$ heads, $\Amat \in \mathbb{R}^{M\times n}$ is the $M$-head attention matrix and $\Zmat\in \mathbb{R}^{M\times d}$ is the resulting sentence representation matrix contains semantics from multiple aspects.

\section{Additional Experimental Details}

For our approach, RBF kernel $\kappa(x,y)=\text{exp}(-\frac{1}{h}\|x-y\|_2^2)$ with the bandwidth $h={med}^2 / \log M$ is used as the kernel function, where $med$ denotes the median of the pairwise distance between current particles. 
The prior distribution of attention parameters is assumed to be uniform. 
We find that adding an repulsive weight before the repulsive term (i.e. the second term in Eq. \ref{svgd2}) in particle-optimization update rules could help adjusting the degree of diversity in attention and achieving better performance. In our experiments, we adopt this trick and use the hyper-parameter $\alpha$ to denote the repulsive weight. 
Since our method only modifies the learning process of attention, all models and settings in our experiments kept the same with the corresponding previous work unless stated otherwise. 

\subsection{Self-attentive Sentence Classification}
\paragraph{Dataset}
Three tasks are conducted on three public sentence classification datasets. \textit{Author profiling} (Age dataset \footnote{https://pan.webis.de/clef16/pan16-web/author-profiling.html}) is to predict the age range of the user by giving their tweets. \textit{Sentiment analysis} (Yelp dataset \footnote{https://www.yelp.com/dataset/download}) is to predict the number of stars the user assigned to by analysis their reviews. \textit{Textual entailment} (SNLI dataset \footnote{https://nlp.stanford.edu/projects/snli/}) is to tell whether the semantics in the two sentences are entailment or contradiction or neutral.
Following \citet{zhouhan}, the train / validate / test split of Age is 68485 / 4000 / 4000, Yelp is 500K / 2000 / 2000, SNLI is 550K / 10K / 10K.

\paragraph{Experimental settings}
We implement the standard multi-head attention model in \citet{zhouhan} following the settings in it except that we use Spacy toolkit \footnote{https://spacy.io/} as the tokenizer and GloVe \footnote{https://nlp.stanford.edu/projects/glove/} (GloVe 840B 300D) as the pre-trained word embedding. 
For repulsive multi-head attention learning, we keep all settings the same with the standard one \cite{zhouhan}. 
Hyper-parameters $\epsilon$ and $\alpha$ in SVGD are selected with grid search. For SPOS, we fix these two hyper-parameters and only tune $\beta$. The selection is based on the performance on the validation data.
We train and evaluate all the models with 10 random seeds and compare their average performance. 
Models are trained on one TITAN Xp GPU. 

\subsection{Transformer-based Neural Translation}
\paragraph{Dataset}
IWSLT14 German-to-English (De-En) dataset contains 153K / 7K / 7K sentence pairs. WMT14 English-to-German (En-De) dataset contains about 4.5 million training sentence pairs and uses newstest2013 dataset as the validation set, newstest2014 dataset as the test set. Data and the processing scripts could be found here \footnote{https://github.com/pytorch/fairseq/tree/v0.6.0/examples/translation}.

\paragraph{Experimental settings}
Our implementation is based on the open-sourced \texttt{fairseq} \footnote{https://github.com/pytorch/fairseq} \cite{fairseq}. We follow the settings in \citet{attention} and have reproduced their reported results.
For the WMT14 dataset, the \texttt{base} Transformer is used, which consists of a 6-layer encoder and a 6-layer decoder. 
The size of the hidden units and embeddings is 512 and the number of heads is 8. The \texttt{big} Transformers has 1024 hidden units and 16 heads, which is listed as a reference. For IWSLT14 dataset, the \texttt{small} setting is used and the number of heads in every layer is set to 4. 
All the configurations are kept the same when applying our method or the regularization method.
In our method, only the training process is changed and SVGD update rule is utilized in our algorithm. The stepsize $\epsilon$ in our method is set to 0.1 and the repulsive term $\alpha$ is set to 0.01. 
The Transformer-\textit{small} model is trained on one TITAN Xp GPU. The Transformer-\textit{base} model is trained on four GTX 1080Ti GPUs.

\paragraph{Additional Results}
To support a fair comparison, we also evaluate the Transformer-base model on WMT14 En-De task. The SACREBLEU score \cite{sacrebleu} with and without our approach
is 27.1 and 26.2\footnote{SacreBLEU hash: \texttt{BLEU+case.mixed+lang.en-de +numrefs.1+smooth.exp+test.wmt14/full+tok.
13a+version.1.4.12}}, respectively.

\subsection{ELECTRA}
\paragraph{Dataset}
Following the official code of \citet{electra}, ELECTRA models are pretrained on the OpenWebTextCorpus  \footnote{https://skylion007.github.io/OpenWebTextCorpus/} dataset, an open source effort to reproduce OpenAI’s WebText dataset. OpenWebTextCorpus containes 38GB of text data from 8,013,769 documents. 
The pretrained model is then finetuned and evaluated on GLUE benchmark \footnote{https://gluebenchmark.com/}. 
GLUE contains a variety of tasks covering textual entailment (RTE and MNLI) question-answer entailment (QNLI),
paraphrase (MRPC), question paraphrase (QQP), textual similarity (STS), sentiment (SST), and linguistic acceptability (CoLA). 
Our evaluation
metrics are Spearman correlation for STS, Matthews correlation for CoLA, and accuracy for the
other GLUE tasks.

\paragraph{Experiment settings}
The ELECTRA-\textit{small} model we implemented follow all official settings \footnote{https://github.com/google-research/electra} except that it is fully-trained on one GTX 1080Ti GPU for 6 days. The ELECTRA-\textit{small} model has 12 layers with 4 heads in every layer's attention.
For our method, the stepsize $\epsilon$ is set to 0.01 and the repulsive term $\alpha$ is set to 0.1. 
The repulsive learning of attention is only applied to the pre-training stage. The fine-tuning remains the same with the original one. 

\subsection{GraphWriter}
\paragraph{Dataset}
Experiments are conducted on the Abstract GENeration DAtaset (AGENDA) \cite{Graphwriter}, a dataset of knowledge graphs paired with scientific abstracts. It consists of 40k paper titles and abstracts from the Semantic Scholar Corpus taken from the proceedings of 12 top AI conferences. We use the standard split of AGENDA dataset in our experiments: 38,720 for training, 1000 for validation, and 1000 for testing.

\paragraph{Experimental settings}
We follow the official settings \footnote{https://github.com/rikdz/GraphWriter} in \citet{Graphwriter} with the encoder containing 6 layers and 4-head graph attention in every layer.
We reproduce their results and keep all settings the same when applying the proposed repulsive attention. 
The SVGD update rule is used in our algorithm and applied to all layers. The stepsize $\epsilon$ is set to 0.1 and the repulsive weight is set to 0.01 in this experiment. The model is trained on one TITAN Xp GPU.

\section{Additional Analysis of Our Approach}
\subsection{Comparison with SGLD}
We also conducted a comparison of our method with Stochastic gradient Langevin dynamics (SGLD) \cite{sgld}, which is also a Bayesian sampling method. Results are in Table \ref{tab:sgld}. Though random noise brought by SGLD might help achieving diversity, it's sub-optimal. Using particle-optimization to add the repulsive term makes it more
effective.

\begin{table}[h]
\small
    \centering
    \begin{tabular}{c c  c  c  c}
    \toprule
    \textbf{Sampling} &&&&\\
    \textbf{Method} &\textbf{Age} & \textbf{Yelp} & \textbf{SNLI} & \textbf{IWSLT14 De-En} \\
    \midrule
      None & 81.47 & 69.3 & 83.79 & 34.4\\
      SGLD & 81.57 & 70.1 & 83.80 & 34.7\\
      SVGD & \bf81.82 & \bf71.2 & \bf84.58 & \bf35.2\\
    \bottomrule
    \end{tabular}
    \caption{Performance of attention models with different sampling methods on four tasks. For Age, Yelp and SNLI sentence classification tasks, the evaluation metric is accuracy (\%). For IWSLT14 De-En translation task, the evaluation metric is BLEU score. "None" here means the standard multi-head attention models.}
    \label{tab:sgld}
\end{table}

\subsection{What Prior to Use?}
In our approach, the repulsiveness is imposed by the inference algorithm (\textit{i.e.} SVGD), not prior. 
To study the impact of different priors, we also tested the Gaussian prior. We found that (see Table \ref{tab:gaussian}) different
priors have little impact on the final results, \textit{i.e.}, there is not a consistent winner for different priors. This suggests that, the prior has little impact on repulsiveness in our framework. But one can still impose prior knowledge of the attention to help our algorithm learn a better attention model. We would like to explore that in future works.
\begin{table}[h]
\small
    \centering
    \begin{tabular}{cccc}
    \toprule
      \textbf{Prior}  &  \textbf{Age} & \textbf{Yelp} & \textbf{SNLI}\\
    \midrule
        Uniform & 81.82 & 71.7 & 84.58 \\
        Gaussian & 81.88 & 71.6 & 84.28 \\
    \bottomrule
    \end{tabular}
    \caption{Performance of our approach with different priors on three sentence classification tasks.The evaluation metric is accuracy (\%)}
    \label{tab:gaussian}
\end{table}

\subsection{Which attention modules to be diversified?}\label{app:Transformer-abl}
\begin{table}[h]
\small
    \centering
    \begin{tabular}{l c c}
        \toprule
        \textbf{Models} & \textbf{BLEU} & \textbf{Time}\\
        \midrule
        MA & 34.4 & 1 \\
        \midrule
        RMA (Q) & 34.7 & 1.06 \\
        RMA (K) & 34.7 &  \\
        RMA (V) & 34.9 &  \\
        \midrule
        RMA (En) & 34.6 & 1.06 \\
        RMA (De) & 34.7 &  \\
        RMA (En-De) & 34.9 &  \\
        \midrule
        RMA (first layer) & \textbf{35.1} & 1.03 \\
        RMA (last layer) & 34.7 &  \\
        \midrule
        RMA (All) & \textbf{35.2} & 1.13\\
       
        \bottomrule
    \end{tabular}
    \caption{Ablation study of Transformer-\texttt{small}-RMA model on IWSLT14 De-En dataset with repulsive multi-head attention applied on different part of the model. En: self-attention in encoder. De: self-attention in decoder. En-De: inter-attention between encoder and decoder. Time: relative training time of every step versus MA.}
    \label{tab:Transformer-abl}
\end{table}
There are three types of attention in Transformer: self-attention in the encoder, self-attention in the decoder, and inter-attention between the encoder and decoder.
We conduct extra experiments on Transformer-\texttt{small} to investigate which attention module benefits most from the repulsiveness. Results are shown in Table \ref{tab:Transformer-abl}. 
We first apply the repulsive attention on each of \{Q,K,V\} parameters in every attention module for all layers. The results indicate that diversifying the $V$-parameter seems to yield better performance. 
We then compare repulsive attention inside the encoder, inside the decoder and between them, respectively. The results show improvement in all cases, and diversifying inter-attention seems to achieve the most benefit. 
Finally, we diversify the attention in different layers of the Transformer. The results suggest that only diversifying the attention in the first layer is able to achieve comparable performance to the case of diversifying all layers, with little computational time increased.

\subsection{Improved Calibration}
\begin{figure*}
\centering
\subfloat[]{{\includegraphics[width=0.3\textwidth]{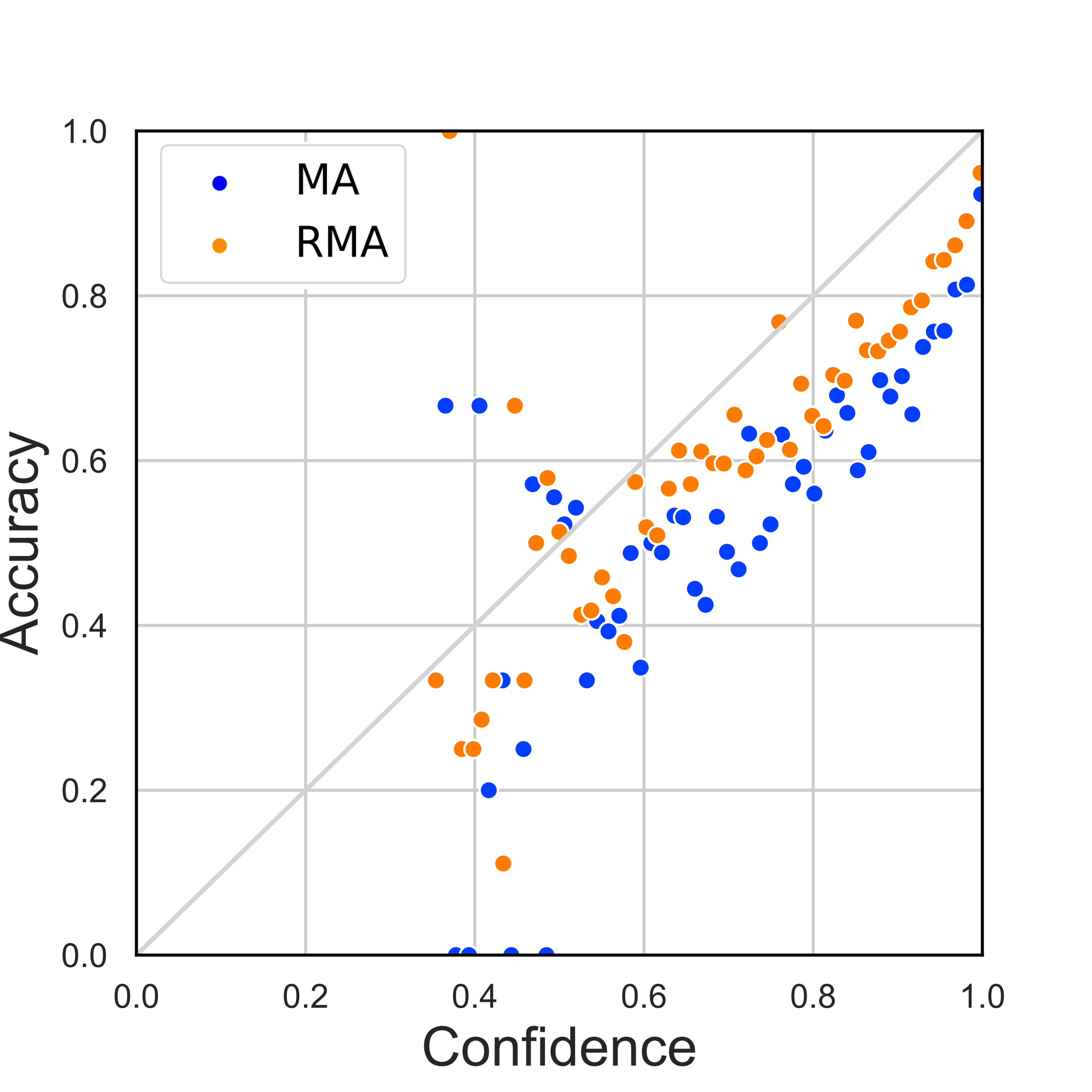}}} 
 \subfloat[]{{\includegraphics[width=0.3\textwidth]{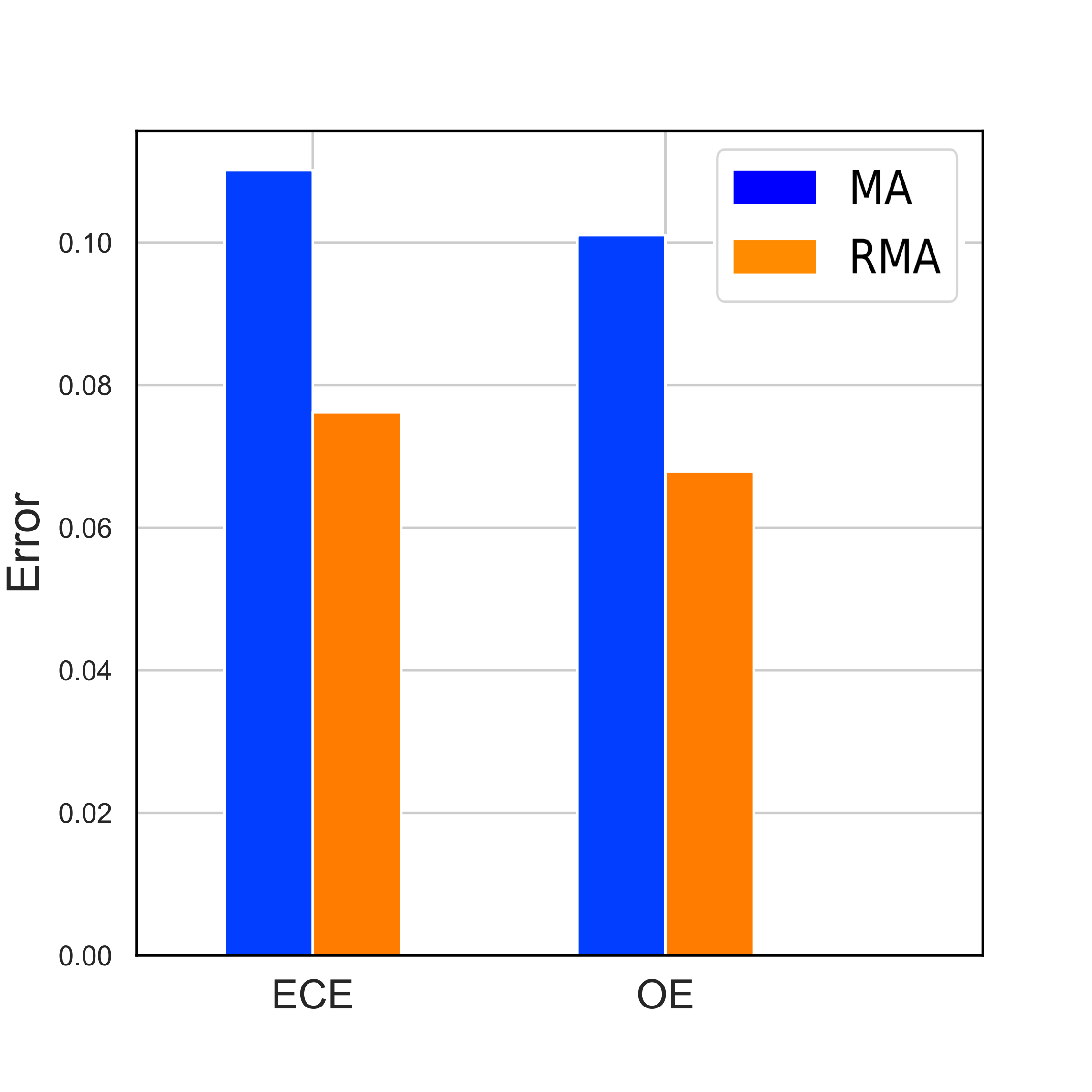}}} \\
    \caption{(a) Scatterplots for accuracy and confidence for SNLI test set. The repulsive case (RMA) is much better calibrated with the points lying closer to the $y = x$ line, while in the standard case (MA), points tend to lie in the overconfident region. (b) Expected Calibration Error (ECE) and Overconfidence Error (OE) for two cases.}
    \label{fig:calibration}
\end{figure*}
A reliable model must not only be accurate, but also indicate when it is likely to get the wrong answer. It means the confidence of a well calibrated model should be indicative of the actual likelihood of correctness. 
Following the calibration metrics in~\citet{calibration} and \citet{uncertainty}, we evaluate the calibration of the model in Figure \ref{fig:calibration}. 
For classifiers, the predicted softmax scores of winning class are represented as the confidence of models. 
Expected Calibration Error (ECE) and Overconfidence Error (OE) are two calibration metrics evaluating the reliability of a model. 
Following \citet{calibration} and \citet{uncertainty}, softmax predictions are grouped into $M$ interval bins of equal size. Let $B_m$ be the set of samples whose prediction scores (the winning softmax score) fall into bin $B_m$. The accuracy and confidence of $B_m$ are defined as 
\begin{align*}
    \text{acc}(B_m) &= \frac{1}{|B_m|} \sum_{i\in B_m}\textbf{1}(\hat{y}_i=y_i)\\
    \text{conf}(B_m) &= \frac{1}{|B_m|} \sum_{i\in B_m}\hat{p}_i
\end{align*}
where $\hat{p}_i$ is the confidence (winning score) of sample $i$. 
The Expected Calibration Error (ECE) is then defined as:
\begin{equation*}
    \text{ECE} = \sum_{m=1}^M \frac{|B_m|}{n}|\text{acc}(B_m) - \text{conf}(B_m)|
\end{equation*}
In high-risk applications, confident but wrong predictions can be especially harmful. 
Overconfidence Error (OE) is defined as follow for this case. 
\begin{align*}
    \text{OE} = &\sum_{m=1}^M \frac{|B_m|}{n}[\text{conf}(B_m) \times \\ &\text{max}(\text{conf}(B_m) - \text{acc}(B_m), 0)] 
\end{align*}
As shown in Figure \ref{fig:calibration}, the standard attention model is prone to be over-confident, meaning that the accuracy is likely to be lower than what is indicated by the predictive score. 
With the proposed repulsive training of attention, the model becomes better calibrated with much lower calibration error and overconfidence error, indicating that our method is beneficial for training more reliable multi-head attention models.

\subsection{Improved Uncertainty Prediction}
\begin{figure}
    \centering
    \includegraphics[width=0.5\textwidth]{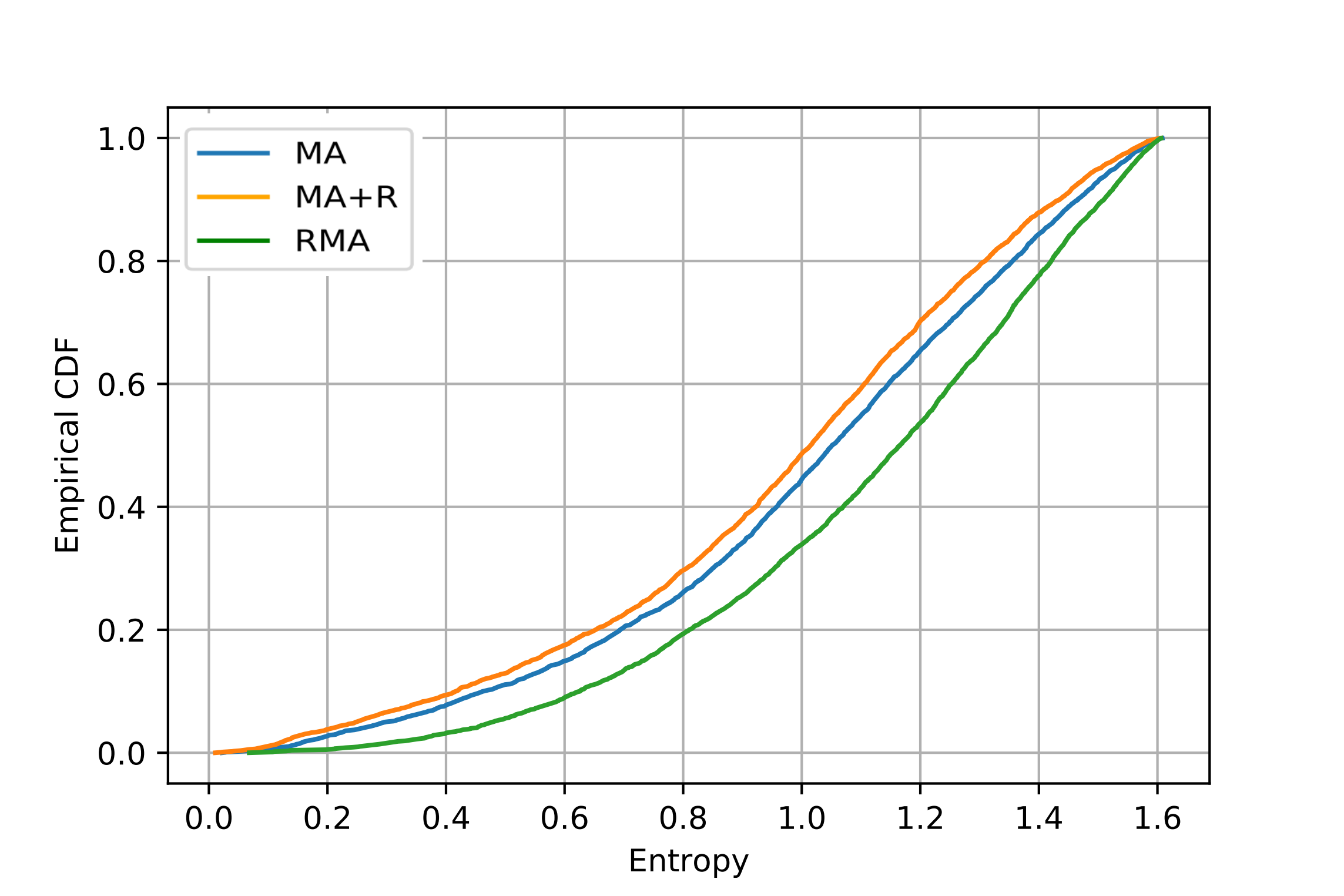}
    \caption{Empirical CDF for the entropy of the predictive distributions on Age dataset from the model trained on Yelp dataset. Curves that are closer to the bottom right part of the plot are preferable, as it denotes that the probability of observing a high confidence prediction is low.}
    \label{fig:uncertainty_entropy}
\end{figure}
We follow \citet{uncertainty_entropy} to evaluate the predictive uncertainty. 
We estimate the entropy of the predictive distributions on Age dataset (out-of-distribution entropy)
from the models trained on Yelp dataset. 
Since we a-priori know that none of the Age classes correspond to a trained class
(they are two different tasks), the ideal predictive distribution is uniform over the Age dataset, i.e. a maximum entropy distribution. 
We plot the empirical CDF for the entropy in Figure \ref{fig:uncertainty_entropy}. 
It shows that the uncertainty estimates from the repulsive multi-head attention model is better than the standard attention and the regularization approach.

\subsection{Discussion with Existing Regularization-based Methods}\label{app:regu}
\begin{table}[]
\small
    \centering
    \begin{tabular}{r|cc}
    \toprule
    \multirow{2}{*}{\bf Methods} & \multicolumn{2}{c}{\bf Acc(\%)}\\
    \cline{2-3}
      & Age & Yelp \\
     \midrule
      cosine similarity regularization & 81.35 & 68.50\\
      $i ->j$   & 81.65 &71.45\\
      $i ->j$, +smooth  & 81.85 & 71.50\\
      \bottomrule
    \end{tabular}
    \caption{Adapt cosine similarity regularization on attention parameters gradually to our framework. Accuracy of the model is evaluated on the test set of Age and Yelp dataset.}
    \label{tab:R_RMA}
\end{table}
Learning diverse attentions has been proposed in~\citet{zhouhan} and \citet{disagree}, with different regularization techniques to enforce repulsiveness. 
In fact, we can show that existing methods are simplified versions of our framework, but with a potential mismatch between their algorithm and the underlying repulsiveness guarantee. To explain this, we follow \citet{disagree} and apply cosine similarity regularizer to the attention parameter $\thetav$. When negative log-likelihood is used as the loss function and the prior of $\thetav^{(i)}$ is assumed to be uniform, the update function $\phi$ becomes: 
\fontsize{10}{11}
\begin{align}
    \phi(\thetav_{\ell}^{(i)}) = -\nabla_{\thetav_{\ell}^{(i)}}U(\thetav_{\ell}^{(i)}) +
    \nabla_{\thetav_{\ell}^{(i)}}\frac{1}{M}\sum_{j=1}^M \frac{\thetav_{\ell}^{(i)}\thetav_{\ell}^{(j)}}{\Vert \thetav_{\ell}^{(i)} \Vert \Vert\thetav_{\ell}^{(j)}\Vert} \notag
\end{align}
One can consider the cosine similarity as a kernel function. Applying this new kernel function to our framework results in the following update:
\fontsize{10}{11}
\begin{align*}\label{eq:reg}
    \phi(\thetav_{\ell}^{(i)}) &= \frac{1}{M} \sum_{j=1}^M-\frac{\thetav_{\ell}^{(i)}\thetav_{\ell}^{(j)}}{\Vert \thetav_{\ell}^{(i)} \Vert \Vert\thetav_{\ell}^{(j)}\Vert} \nabla_{\thetav_{\ell}^{(j)}}U(\thetav_{\ell}^{(j)})   \\
    &+ \frac{1}{M} \sum_{j=1}^M\nabla_{\thetav_{\ell}^{(j)}}\frac{\thetav_{\ell}^{(i)}\thetav_{\ell}^{(j)}}{\Vert \thetav_{\ell}^{(i)} \Vert \Vert\thetav_{\ell}^{(j)}\Vert} 
\end{align*}
It is clear that our method \eqref{eq:reg} reduces to the regularizing method when 1) removing the smoothing term $\frac{\thetav_{\ell}^{(i)}\thetav_{\ell}^{(j)}}{\Vert \thetav_{\ell}^{(i)} \Vert \Vert\thetav_{\ell}^{(j)}\Vert}$ for $j\neq i$; and 2) replacing the derivative $\nabla_{\thetav_{\ell}^{(j)}}$ with $\nabla_{\thetav_{\ell}^{(i)}}$ in the repulsive term. 
For this reason, we argue that the regularization method lacks of a formal guarantee on the repulsiveness.

To show this, we adapt the cosine similarity regularization on attention parameters gradually to our framework by 1) replacing the derivative $\nabla_{\thetav_{\ell}^{(i)}}$ with $\nabla_{\thetav_{\ell}^{(j)}}$ in the repulsive term; 
2) adding the smoothing term $\kappa(\thetav_{\ell}^{(j)}, \thetav_{\ell}^{(i)})$. 
Table \ref{tab:R_RMA} shows that both the smoothing and the corrected gradient lead to performance improvement over the standard regularization methods.

\newpage
\onecolumn
\section{Visualization of Multi-head Attention in Sentence Classification Task} \label{app:vis}
\begin{figure}[h]
    \centering
    \subfloat[detailed attentions of 30 heads]{\includegraphics[width=0.88\textwidth]{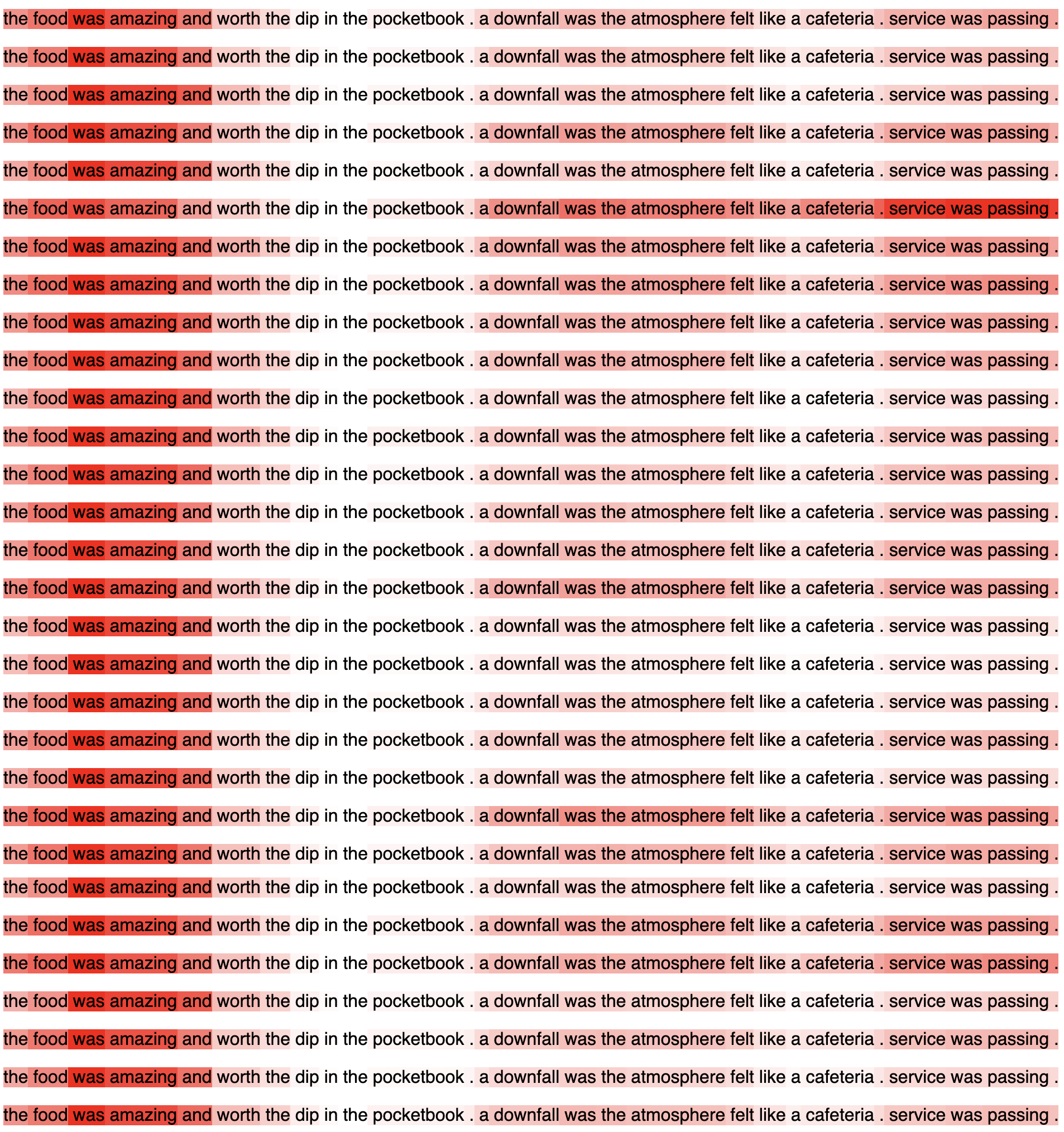}} \\
    \subfloat[overall attention]{\includegraphics[width=0.88\textwidth]{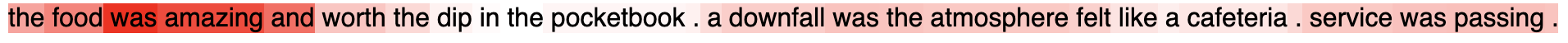}} 
    \caption{Attention heatmaps for the standard multi-head attention model}
    \label{fig:my_label}
\end{figure}

\begin{figure}
    \centering
    \subfloat[detailed attentions of 30 heads]{\includegraphics[width=0.88\textwidth]{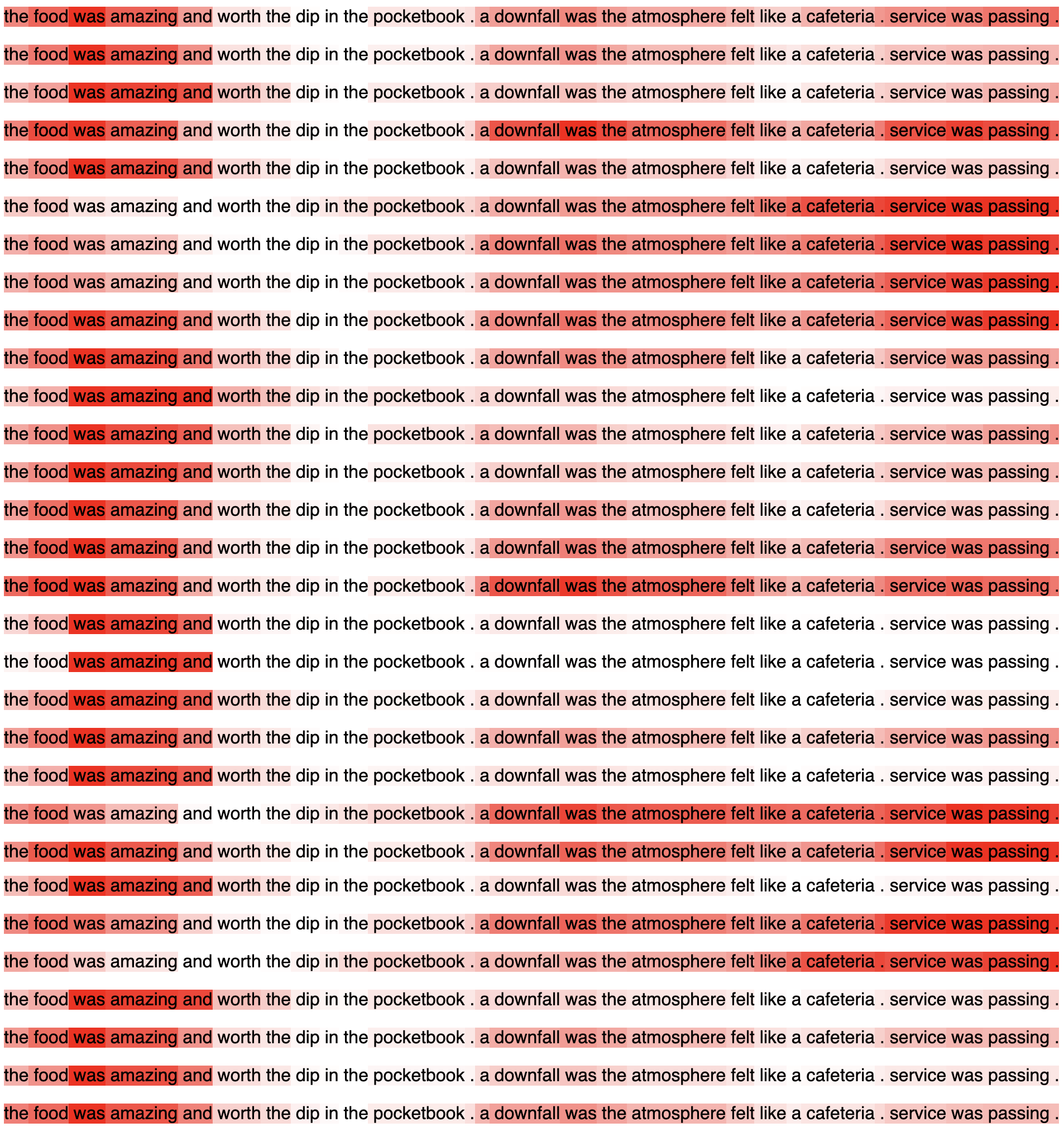}} \\
    \subfloat[overall attention]{\includegraphics[width=0.88\textwidth]{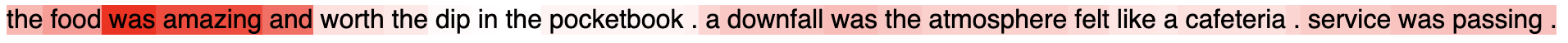}} 
    \caption{Attention heatmaps for the standard multi-head attention model trained with Frobenius regularization}
    \label{fig:my_label}
\end{figure}

\begin{figure}
    \centering
    \subfloat[detailed attentions of 30 heads]{\includegraphics[width=0.88\textwidth]{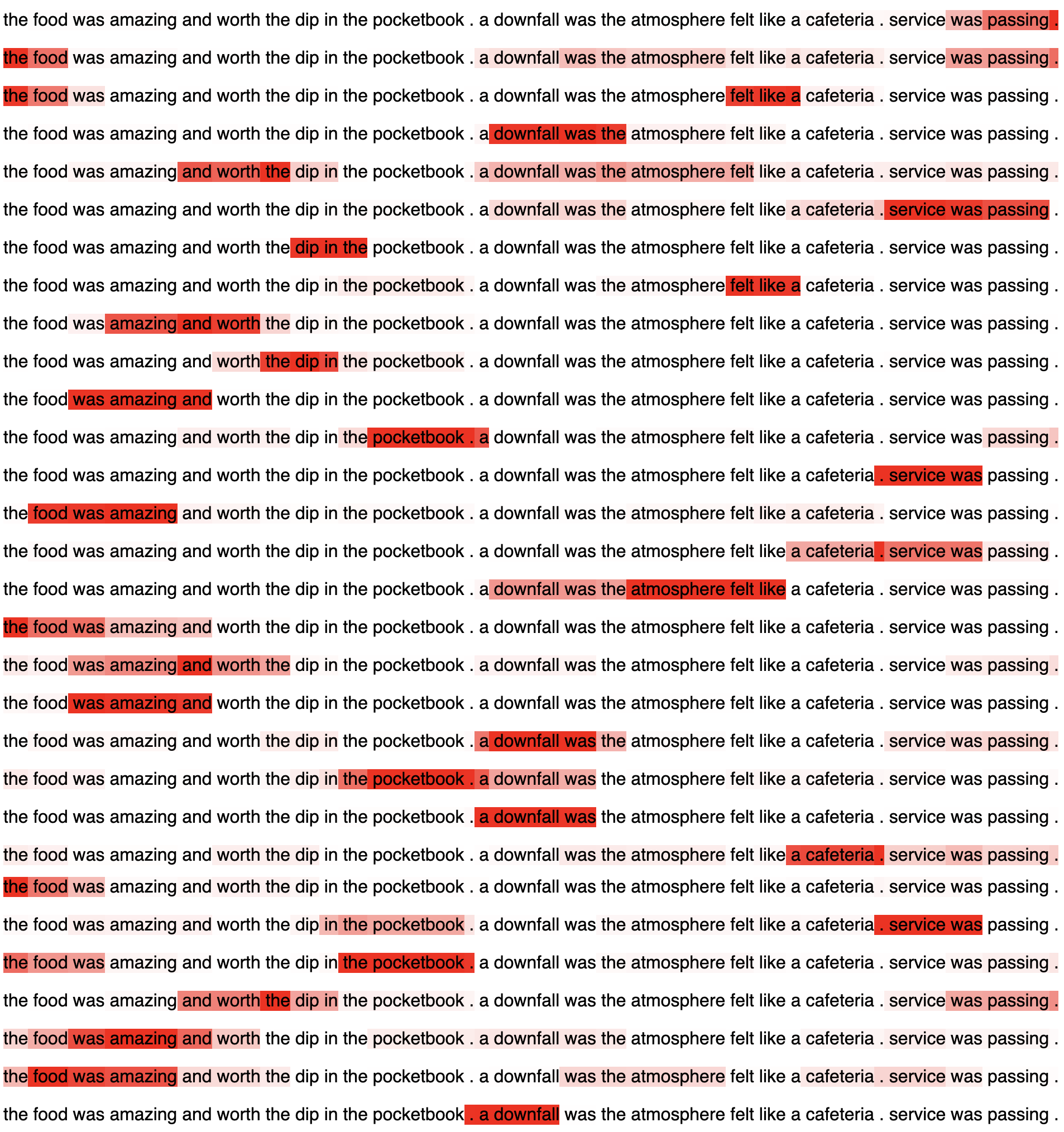}} \\
    \subfloat[overall attention]{\includegraphics[width=0.88\textwidth]{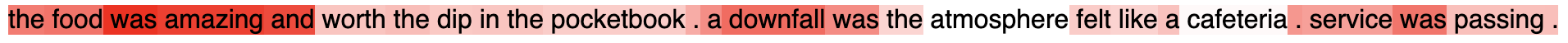}} 
    \caption{Attention heatmaps for the repulsive multi-head attention model}
    \label{fig:my_label}
\end{figure}

\newpage
\onecolumn
\section{A Generation Sample of Graph-to-Text Task} \label{app:GW}
\vspace{2cm}

\begin{figure*}[h]
    \centering
    \includegraphics[width =\textwidth]{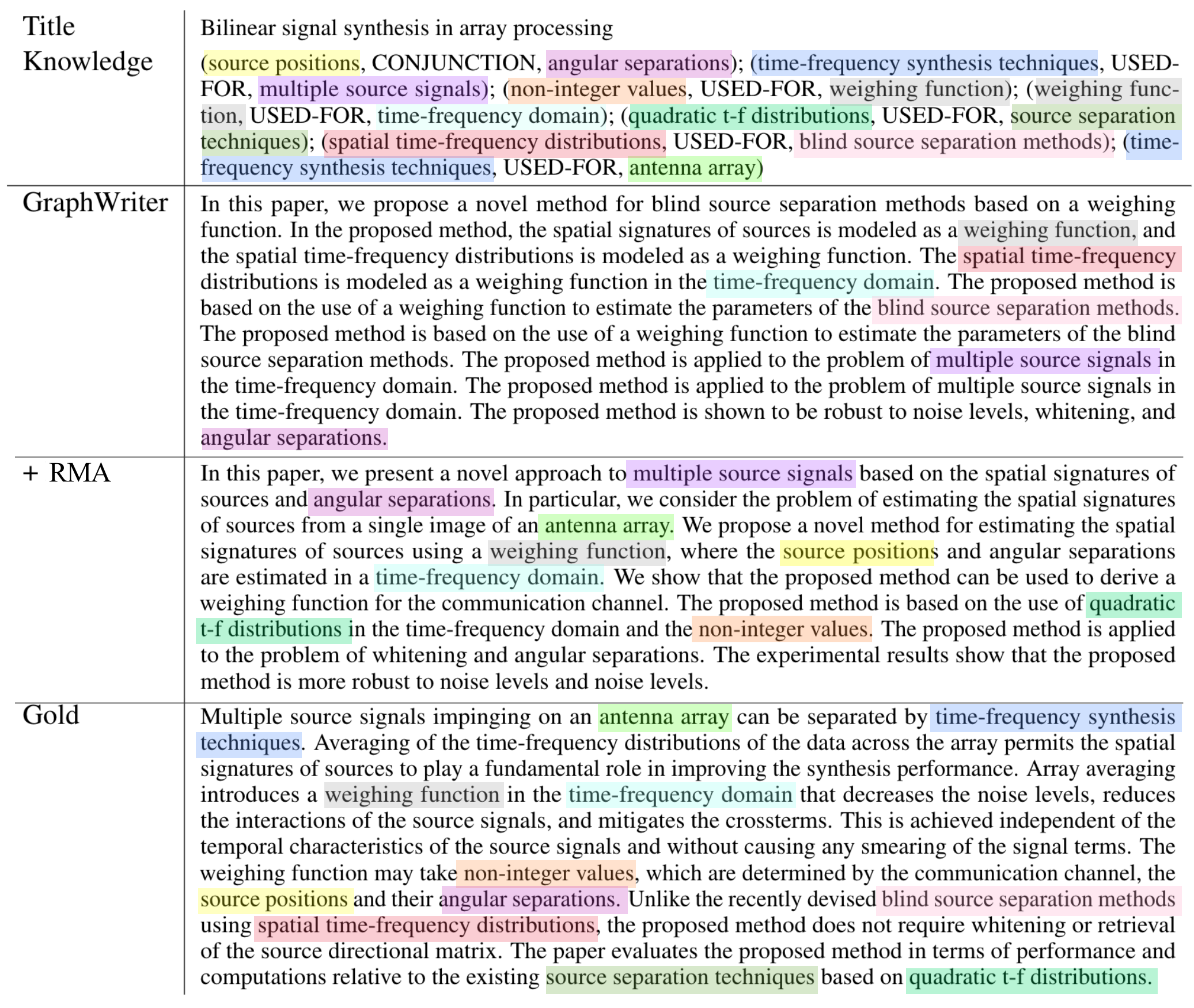}
    \caption{Example outputs of various systems versus Gold}
    \label{tab:GW_gen_example_1}
\end{figure*}